\documentclass[sigconf]{acmart}

\usepackage{mdframed}
\usepackage{amsfonts}
\usepackage{dsfont}
\usepackage{multirow}
\usepackage{algorithm}
\usepackage{algpseudocode}
\usepackage{pgfplots}
\usepackage{caption}
\usepackage{subcaption}
\usepackage{tikz}
\usepackage{listings}
\usepackage{xcolor}
\usepackage{tabularx}

\usetikzlibrary{patterns}
\pgfplotsset{compat=1.17}

\definecolor{boxbg}{RGB}{248,248,248}
\definecolor{boxframe}{RGB}{180,180,180}
\definecolor{hlgreen}{RGB}{220,245,220}
\definecolor{hlred}{RGB}{250,220,220}
\newcommand{\hlg}[1]{\colorbox{hlgreen}{#1}}
\newcommand{\hlr}[1]{\colorbox{hlred}{#1}}
\definecolor{softgreen}{rgb}{0.13, 0.6, 0.13}
\definecolor{softred}{rgb}{0.8, 0.25, 0.25}
\definecolor{colOne}{RGB}{255,242,204}
\definecolor{colTwo}{RGB}{255,217,102}
\definecolor{colThree}{RGB}{230,159,80}

\newcommand{\pos}[1]{\textcolor{softgreen}{+#1}}
\newcommand{\negnum}[1]{\textcolor{softred}{#1}}
\newcommand{\stitle}[1]{\par\vspace{1.2ex}\noindent\textbf{#1}}

\lstdefinestyle{sqls}{
    language=SQL,
    basicstyle=\ttfamily\small,
    keywordstyle=\ttfamily\small\bfseries, 
    stringstyle=\ttfamily\small,        
    commentstyle=\ttfamily\small\itshape, 
    showstringspaces=false,
    breaklines=true,
    frame=none,
    numbers=none,
    escapeinside={(*@}{@*)},
    morekeywords={SELECT, FROM, WHERE, JOIN, ON, GROUP, BY, ORDER, HAVING, DISTINCT, AS, AND, OR, NOT, NULL, IS, IN, LIKE, BETWEEN, LIMIT, OFFSET, INNER, LEFT, RIGHT, OUTER, UNION, INTERSECT, EXCEPT, COUNT, SUM, AVG, MAX, MIN, ROUND, CAST}
}

\AtBeginDocument{%
  }

\copyrightyear{2026}
\acmYear{2026}
\setcopyright{cc}
\setcctype{by}
\acmConference[KDD 2026] {Proceedings of the 32nd ACM SIGKDD Conference on Knowledge Discovery and Data Mining V .2}{August 9--13, 2026}{Jeju Island, Republic of Korea.}
\acmBooktitle{Proceedings of the 32nd ACM SIGKDD Conference on Knowledge Discovery and Data Mining V .2 (KDD 2026), August 9--13, 2026, Jeju Island, Republic of Korea}
\acmISBN{979-8-4007-2259-2/2026/08}
\acmDOI{10.1145/3770855.3818164}
\begin{document}

\title{\textsc{ErrorLLM}: Modeling SQL Errors for Text-to-SQL Refinement}

\author{Zijin Hong}
\affiliation{%
  \institution{The Hong Kong Polytechnic University}
  \city{Kowloon}
  \country{Hong Kong}
}
\email{zijin.hong@connect.polyu.hk}

\author{Hao Chen}
\affiliation{%
  \institution{City University of Macau}
  \city{Taipa}
  \country{Macau}}
\email{sundaychenhao@gmail.com}

\author{Zheng Yuan}
\affiliation{%
  \institution{The Hong Kong Polytechnic University}
  \city{Kowloon}
  \country{Hong Kong}
}
\email{yzheng.yuan@connect.polyu.hk}

\author{Qinggang Zhang}
\authornote{Corresponding author.}
\affiliation{%
  \institution{Jilin University}
  \city{Changchun}
  \country{China}
}
\email{qinggangzhang@jlu.edu.cn}

\author{Luyao Zhuang}
\affiliation{%
  \institution{The Hong Kong Polytechnic University}
  \city{Kowloon}
  \country{Hong Kong}
}
\email{luyao.zhuang@connect.polyu.hk}

\author{Qing Liao}
\affiliation{%
  \institution{Harbin Institute of Technology (Shenzhen)}
  \city{Shenzhen}
  \country{China}
}
\email{liaoqing@hit.edu.cn}

\author{Feiran Huang}
\affiliation{%
  \institution{Beihang University}
  \city{Beijing}
  \country{China}
}
\email{huangfr@buaa.edu.cn}

\author{Yangqiu Song}
\affiliation{%
  \institution{The Hong Kong University of Science and Technology}
  \city{New Territories}
  \country{Hong Kong}
}
\email{yqsong@cse.ust.hk}

\author{Xiao Huang}
\affiliation{%
  \institution{The Hong Kong Polytechnic University}
  \city{Kowloon}
  \country{Hong Kong}
}
\email{xiaohuang@comp.polyu.edu.hk}

\renewcommand{\shortauthors}{Zijin Hong et al.}

\begin{abstract}
    Despite the remarkable performance of large language models (LLMs) in text-to-SQL, correctly producing SQL queries remains challenging during initial generation.
    The SQL refinement task is subsequently introduced to correct syntactic and semantic errors.
    However, existing paradigms face two major limitations:
    (i) self-debugging becomes increasingly ineffective as modern LLMs rarely produce explicit execution errors;
    (ii) self-correction exhibits low detection precision due to the lack of explicit error modeling grounded in the question and schema, and suffers from severe hallucination that frequently corrupts correct SQLs.
    In this paper, we propose \textbf{\textsc{ErrorLLM}}, a framework that explicitly models text-to-SQL \textbf{\textsc{Error}}s within a dedicated \textbf{LLM} for text-to-SQL refinement.
    Specifically, we represent question and schema as structural features, employ static detection to identify execution failures and surface mismatches, and extend \textsc{ErrorLLM}'s semantic space with dedicated error tokens that capture categorized implicit semantic error types.
    Through a well-designed training strategy, we explicitly model these errors with structural representations, enabling the LLM to detect complex implicit errors by predicting dedicated error tokens.
    Guided by the detected errors, we perform error-guided refinement on the SQL structure by prompting LLMs.
    Extensive experiments demonstrate that \textsc{ErrorLLM} achieves the most significant improvements over backbone initial generation.
    Further analysis reveals that detection quality directly determines refinement effectiveness, which \textsc{ErrorLLM} achieves through a high detection F1 score.
    The corresponding code of \textsc{ErrorLLM} is released for further research\footnote{\href{https://github.com/DEEP-PolyU/ErrorLLM}{https://github.com/DEEP-PolyU/ErrorLLM}}.
\end{abstract}




\maketitle

\section{Introduction}

\begin{figure}[!t]
    \centering
    \includegraphics[width=\columnwidth]{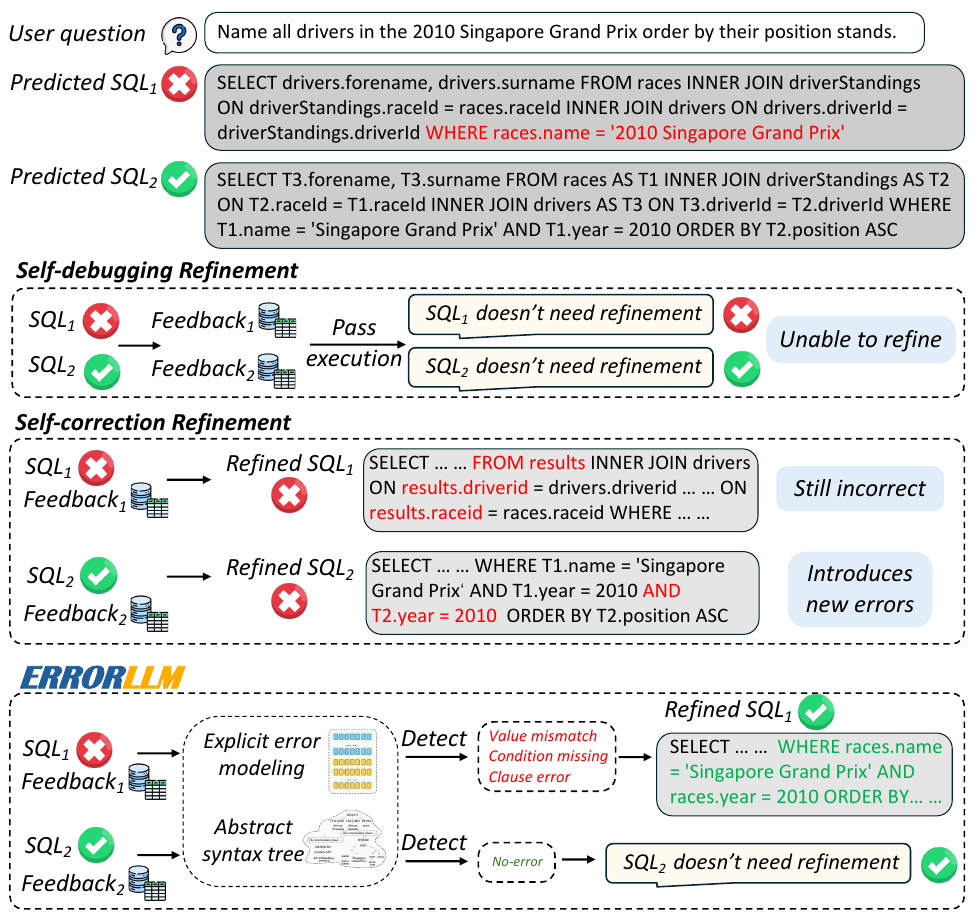} 
    \vspace{-6mm}
    \caption{Different paradigms of text-to-SQL refinement.
    Self-debugging misses the incorrect SQL.
    Self-correction conducts correction on both SQLs but cannot successfully fix the error.}
    \label{figure:introduction}
    \vspace{-5mm}
\end{figure}

Text-to-SQL generation aims to translate user questions (natural language questions) into executable SQL queries~\cite{hong2025next}.
As the crossing point intersecting natural language understanding and database systems, text-to-SQL has been a long-standing research question~\cite{liu2025survey}.
Existing LLM-based text-to-SQL methods often struggle to understand complex questions, database schemas, and the precise structural syntax constraints required during initial SQL generation~\cite{shute2024sql}.
In this case, text-to-SQL refinement becomes a critical and almost compulsory module that refines the generated SQL queries to improve accuracy~\cite{liu2025nl2sqlbug}.
Recent advanced text-to-SQL implementations have carefully studied and designed specific modules to refine the initially generated SQL queries~\cite{wang2024macsql}, which can be categorized into two mainstream paradigms. \textbf{Self-debugging}~\cite{chen2024teach}: LLMs interactively refine their generated SQLs according to execution results that. \textbf{Self-correction}~\cite{pourreza2023dinsql}: LLMs are triggered by elaborated prompt engineering and workflows to semantically analyze the generated SQLs.
Figure~\ref{figure:introduction} provides an intuitive example of both paradigms.
Beyond integrating these modules into complete text-to-SQL workflows, the significance of the text-to-SQL refinement task has raised growing interest in the community towards designing specialized approaches for SQL refinement~\cite{cen2025sqlfixagent,askari2025magic,qu2025share}.

However, recent studies reveal significant limitations in current text-to-SQL refinement modules and methods~\cite{qu2025share,xu2025ts}.
As the most intuitive paradigm, self-debugging approaches refine SQLs using execution feedback as guidance.
Nevertheless, with the rapid development of LLMs, the backbone SQL generation capability has been largely improved~\cite{liu2025nl2sqlbug}.
In our pilot experiments, the syntax errors that raise execution warnings occupy only 3\% of all incorrect SQLs; the remaining SQLs do not raise explicit execution errors and thus cannot be refined by self-debugging paradigms.
For self-correction, this paradigm relies on LLMs' internal reasoning to identify and correct errors without external feedback.
However, without explicit modeling of SQL errors grounded in the question and schema, LLMs inherently struggle to self-assess their own declarative SQL outputs~\cite{qu2025share,xia2024r3}, resulting in low detection precision on incorrect SQLs.
More critically, this low-precision detection leads to severe hallucination during the correction phase: when LLMs are prompted to correct a SQL that is already correct, they tend to comply with the correction instruction regardless, modifying the SQL and producing an erroneous result~\cite{ning2024insights,li2024dawn}.
We refer to this phenomenon as \emph{corruption}, where originally correct SQLs are corrupted into incorrect ones~\cite{gong2025sqlens}.
As evidenced by our experiments, self-correction~\cite{talaei2024chess,pourreza2024chase} achieves a commendable fixed rate but low overall improvement because the substantial number of false-positive detections triggers widespread corruption that offsets successful corrections.

The limitations of both paradigms point to a common challenge: effective text-to-SQL refinement fundamentally requires accurate SQL error detection, yet neither self-debugging nor self-correction provides it~\cite{hong2025next}.
Self-debugging relies on execution signals that cover only a small fraction of errors, while self-correction attempts detection and correction simultaneously in a black-box manner.
We propose that a dedicated \textbf{SQL error detection} module that explicitly identifies error types grounded in the question and schema is the key to enabling effective \textbf{text-to-SQL refinement}.

In this paper, we propose \textbf{\textsc{ErrorLLM}}, a framework that explicitly models text-to-SQL errors within dedicated error tokens for text-to-SQL refinement.
Specifically, we first represent the user question and database schema as structural features, including a question-schema structure and abstract syntax trees for SQL.
For SQL error detection, we employ a two-stage approach: static superficial detection first identifies execution failures and surface mismatches through deterministic rules; we then extend \textsc{ErrorLLM}'s semantic space with dedicated error tokens that represent categorized implicit semantic error types.
Through a well-designed training strategy, we explicitly model these errors that enabling the LLM to effectively detect complex implicit errors by predicting dedicated error tokens.
For text-to-SQL refinement, guided by the detected errors, we perform error localization and analysis to pinpoint error-relevant AST nodes and schema elements, and then conduct priority-ordered error-guided refinement by prompting LLMs.
Extensive experiments on \textsc{BIRD} and \textsc{Spider} benchmarks demonstrate that \textsc{ErrorLLM} achieves the most significant improvements over backbone LLMs, and further analysis on the \textsc{NL2SQL-Bugs} benchmark verifies the effectiveness of fine-grained error type detection. Our contributions are summarized as follows:
\begin{itemize}
    \item We propose \textsc{ErrorLLM}, a framework that explicitly models text-to-SQL errors by extending the LLM's semantic space with dedicated error tokens. Each token corresponds to a specific error type, enabling precise SQL error detection.
    \item We design a comprehensive pipeline consisting of SQL error detection and error-guided text-to-SQL refinement. The detection stage combines static superficial detection with LLM-based semantic detection; the refinement stage leverages error localization, analysis, and priority-ordered context to guide LLMs in producing targeted corrections.
    \item Extensive experiments demonstrate that \textsc{ErrorLLM} achieves the most significant improvements over backbone LLMs on \textsc{BIRD} and \textsc{Spider} benchmarks. Further analysis confirms that the quality of SQL error detection directly determines the effectiveness of text-to-SQL refinement.
\end{itemize}

\section{Preliminaries}

In this section, we first define the tasks, and then introduce the structural representations used as input to our framework.
\subsection{Text-to-SQL Generation}
Given a user question $q$ and a corresponding database $\mathcal{D}$, we define the database schema $\mathcal{S}$ as a pair $(\mathcal{T},\,\mathcal{C})$, where $\mathcal{T}$ is the set of tables and $\mathcal{C}$ is the set of columns. 
The text-to-SQL task aims to learn a mapping $f_{\textsc{t2S}}$ to convert the question into SQL such that:
\begin{equation}
    f_{\textsc{t2S}}: (q,\,\mathcal{S}) \to y,
\end{equation}
where $y$ is the ground-truth SQL query corresponding to the natural language question. Specifically, for LLM-based text-to-SQL~\cite{hong2025next}:
\begin{equation}
    \hat{y} = \operatorname{LLM}(\mathcal{I},\,q,\,\mathcal{S}),
\end{equation}
where $\operatorname{LLM}$ denote a large language model; $\mathcal{I}$ is the set of text-to-SQL instructions; $\hat{y}$ is the predicted SQL query for $q$.

\subsection{Text-to-SQL Refinement}
Following recent advances in the community, we formalize text-to-SQL refinement and introduce two dominant paradigms: self-debugging~\cite{chen2024teach} and self-correction~\cite{pourreza2023dinsql}.
Given a predicted SQL $\hat{y}$, the text-to-SQL refinement is to build a mapping $f_{\textsc{Ref}}$ such that:
\begin{equation}
\label{equation:refinement}
    f_{\textsc{Ref}}: \hat{y} \to \tilde{y},\, \operatorname{\textsc{Exec}}(\tilde{y},\,\mathcal{D}) = \operatorname{\textsc{Exec}}(y,\,\mathcal{D}), 
\end{equation}
where $\operatorname{\textsc{Exec}(\cdot,\,\cdot)}$ is the execution operator that produce execution result/error; $\tilde{y}$ denote the refined SQL.
The task is to produce an execution-equivalent SQL query to ground-truth SQL $y$.

\stitle{Self-debugging.}
The self-debugging paradigm executes the predicted $\hat{y}$ against $\mathcal{D}$ and leverages the execution feedback to iteratively refine the SQL via an LLM:
\begin{equation}
\tilde{y}^{(k+1)} = \operatorname{\textsc{DebLLM}}\!\left(q,\mathcal{S},\hat{y}^{(k)}, \operatorname{\textsc{Exec}}(\hat{y}^{(k)}, \mathcal{D})\right),
\end{equation}
where $\operatorname{\textsc{DebLLM}}$ denotes a self-debugging LLM. 
Multi-round self-debugging incorporating interaction $k$ is widely used~\cite{wang2024macsql}.

\stitle{Self-correction.}
The self-correction paradigm involves refining the predicted SQL by LLMs' internal reasoning process:
\begin{equation}
\tilde{y}^{(k+1)} = \operatorname{\textsc{CorLLM}}\!\left(q,\,\mathcal{S},\,\hat{y}^{(k)},\,\tilde{\mathcal{I}}\right),
\end{equation}
where $\operatorname{\textsc{CorLLM}}$ denotes a self-correction LLM and $\tilde{\mathcal{I}}$ denotes correction guidelines defined by the respective framework~\cite{pourreza2023dinsql}.

\subsection{Structural Representations in Text-to-SQL}
\label{section:structure}
Recent studies indicate that LLMs struggle significantly to correct their own declarative SQL outputs~\cite{qu2025share,zhang2024sgusql}.
We therefore employ structural representations, rather than flat textual input, to capture the complex relations involved in our framework~\cite{zhou2026graph}.

\stitle{Question-schema Structure.} 
We first build a structural representation of the question and schema to support the following process.
To bridge the gap between $q$ and $\mathcal{S}$, we construct a question-schema structure (QSS), denoted as $\mathcal{G} = (\mathcal{V}_{\mathcal{G}},\,\mathcal{E}_{\mathcal{G}})$. 
Concretely, $\mathcal{G}$ unifies two key structures: (i) the \textbf{database structure ($R_{\mathcal{S}}$)} encoding intrinsic relations in the schema (e.g., table/column relations, primary keys, foreign keys), and (ii) the \textbf{linking structure ($R_\textsc{Link}$)} aligning question phrases with relevant tables/columns~\cite{zhang2024sgusql}. 
\begin{equation}
\label{equation:qss}
\mathcal{V}_{\mathcal{G}} = \mathcal{V}_q \cup \mathcal{V}_{\mathcal{S}};\;\mathcal{E}_{\mathcal{G}} = \mathcal{E}_{\mathcal{S}} \cup \mathcal{E}_{\textsc{Link}},
\end{equation}
where $\mathcal{V}_q$ denotes nodes corresponding to tokens/phrases in $q$, and $\mathcal{V}_{\mathcal{S}}$ denotes schema nodes (tables and columns) in $\mathcal{S}$. 

\noindent \textbf{(i) Database structure ($R_\mathcal{S}$).} 
We construct a schema structure $\mathcal{G}_{\mathcal{S}}=(\mathcal{V}_{\mathcal{S}},\,\mathcal{E}_{\mathcal{S}})$ where $\mathcal{E}_{\mathcal{S}}$ contains structural relations within $\mathcal{S}$, including primary- and foreign-key relations between columns~\cite{cao2023heterogeneous}.

\noindent \textbf{(ii) Linking structure ($R_{\textsc{Link}}$).} To connect $\mathcal{G}_q$ and $\mathcal{G}_\mathcal{S}$, we establish cross edges $\mathcal{E}_{\textsc{Link}}$ that link question nodes to candidate schema nodes that map question phrases to their relevant tables or columns. 
The linking structure $R_{L}$ can be constructed using existing graph-based algorithms~\cite{wang2020rat,cao2023astormer,cao2023heterogeneous,zhang2024sgusql}, or obtained directly from a schema linking function $f_{\textsc{SL}}:\mathcal{V}_q \rightarrow 2^{\mathcal{V}_\mathcal{S}}$ from existing methods~\cite{pourreza2023dinsql,yuan2025knapsack}.

\begin{figure*}[!t]
\centering
    \includegraphics[width=\linewidth]{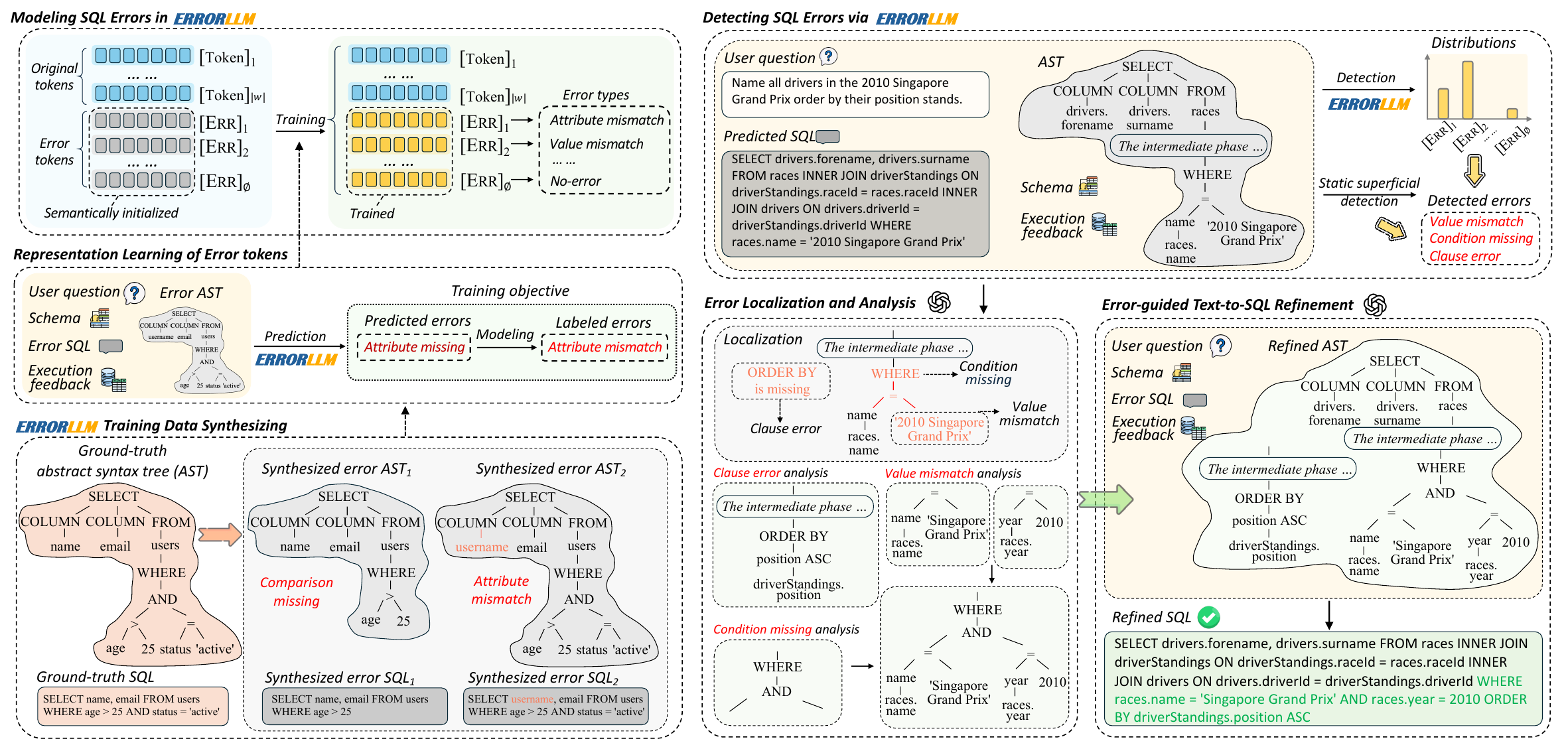}
    \vspace{-6mm}
    \caption{An illustration of the modeling the SQL error in \textsc{ErrorLLM} and the overview of SQL error detection and error-guided text-to-SQL refinement process. 
    The example is selected from the \textsc{BIRD}~\cite{li2023bird} development set, and refined by our workflow.}
    \vspace{-3mm}
\label{figure:overview}
\end{figure*}

\stitle{Abstract Syntax Tree for SQL.} 
Instead of prompting LLMs to directly refine the complete declarative SQL, we convert the predicted SQL into an abstract syntax tree (AST) $\tau_{\hat{y}}$ and conduct error-guided refinement on this structured representation. 
We provide an AST example in Figure~\ref{figure:overview}, where each node of $\tau_{\hat{y}}$ corresponds to a syntactic unit in $\hat{y}$, such as a SQL clause, operator or schema element. Formally, we define the AST as a tree:
\begin{equation}
    \tau_{\hat{y}} = (\mathcal{V}_{\tau_{\hat{y}}},\,\mathcal{E}_{\tau_{\hat{y}}}),\,v \in \mathcal{V}_{\tau_{\hat{y}}},
\end{equation}
where $v$ is the node.
To explicitly connect the structural representation with the declarative SQL query $\hat{y} = (\hat{y}_{1}, \hat{y}_{2}, \ldots, \hat{y}_{|\hat{y}|})$, we define an alignment function:
\begin{equation}
\label{equation:alignment}
    \pi: \mathcal{V}_{\tau_{\hat{y}}} \rightarrow 2^{\{1,\dots,|\hat{y}|\}},\, \pi(v) \in \{1,\dots,|\hat{y}|\},
\end{equation}
where $\pi(v)$ returns the token indices in $\hat{y}$ that correspond to the syntactic unit represented by node $v$.
This alignment ensures that error localization results in $\tau_{\hat{y}}$ can be deterministically mapped back to token spans in the declarative SQL query.

\section{Methods}

\label{section:methods}
In this section, we present the proposed methods from a workflow perspective and detail the training process of \textsc{ErrorLLM}.

\subsection{Modeling SQL Error in \textsc{ErrorLLM}}
\label{section:modeling}
To explicitly model SQL errors with structural representations, we map descriptive error representations into the LLM's semantic space via dedicated tokens.
With this modeling, the LLM can explicitly associate specific error types with incorrect SQLs given the structural context defined above.
By learning these dedicated representations, the model decouples the error detection capability, while preserving its foundational text-to-SQL capabilities~\cite{li2025jarvis}.

\stitle{Error Token Initialization in \textsc{ErrorLLM}.} In detail, we extend the LLM's semantic space by introducing dedicated error tokens into the vocabulary, decoupling error-specific representations from linguistic ones.
Denoting the detection model as $\operatorname{\textsc{ErrorLLM}}$, we extend the original LLM's vocabulary $\mathcal{W}$ with a reserved set of $N$ indexed error tokens $[\textsc{Err}]$ and a special null token $[\textsc{Err}]_{\varnothing}$:
\begin{equation}
    \mathcal{W}' = \mathcal{W} \cup \{[\textsc{Err}]_{i}\}_{i=1}^{N} \cup \{[\textsc{Err}]_{\varnothing}\},
\end{equation}
where $N$ denotes the maximum number of supported error types; $\mathcal{W}'$ denotes the extended vocabulary.
For the error detection task, we have a pre-defined error taxonomy $\Lambda$ as the detection target.
We establish a mapping where the ``no-error'' case $\lambda_{\varnothing}$ maps to $[\textsc{Err}]_{\varnothing}$, and each active error type $\lambda$ maps to a unique indexed token:
\begin{equation}
\label{equation:error-assign}
    \lambda_{\varnothing} \mapsto [\textsc{Err}]_{\varnothing};\; \lambda_{i} \mapsto [\textsc{Err}]_{i},\,i \in \{1,\dots,|\Lambda|\}.
\end{equation}
To ensure scalability to future error types, we set the vocabulary capacity such that $N \geq |\Lambda|$. 
Before training, the embeddings of error tokens are initialized by the averaging of related semantics:
\begin{equation}
\label{equation:initialization}
    \mathbf{e}_{[\textsc{Err}]_{i}} := \frac{1}{|\mathcal{W}_{\lambda_{i}}|} \sum_{w \in \mathcal{W}_{\lambda_{i}}} \mathbf{e}_w,\,i \in \{1,\dots,|\Lambda|\}
\end{equation}
where $\mathcal{W}_{\lambda_{i}}$ is a set of semantically related words derived from the description of error type $\lambda$, and $\mathbf{e}_w$ denotes the pre-trained embedding of token $w \in \mathcal{W}$.
As an example, the related words of ``table redundancy'' include ``extra'', ``redundant'', and ``unnecessary columns''; the $[\textsc{Err}]_{\varnothing}$ is initialized by averaging the embeddings of ``correct'' and ``SQL''.
Compared with random initialization, Equation~\ref{equation:initialization} provides a more effective starting point for training.

\stitle{\textsc{ErrorLLM} Training Data Synthesis.}
Training the error tokens amounts to learning representations that align text-to-SQL errors with the structural representation.
Inspired by previous studies~\cite{li2024codes,li2025omnisql,qu2025share}, we synthesize the training data of error tokens by the rule-based perturbation operators and the LLM-assisted error injections. 

\noindent \textbf{(i) Rule-based perturbation operators.}
For each error type $\lambda_i \in \Lambda$, we define a corresponding AST-level perturbation operator $\operatorname{\textsc{Rule}}_{\lambda_i}$ that transforms the ground-truth AST $\tau_{y}$ into an erroneous variant $\tau_{y'}$ specific to the error type $\lambda_{i}$:
\begin{equation}
\label{equation:perturbation}
    \tau_{y'} = \operatorname{\textsc{Rule}}_{\lambda_{i}}(\tau_{y}, \mathcal{S}),\, \operatorname{\textsc{Exec}}(y', \mathcal{D}) \neq \operatorname{\textsc{Exec}}(y, \mathcal{D}).
\end{equation}
Each operator performs targeted structural modifications guided by $\mathcal{S}$, the flat SQLs $y'$ and $y$ are obtained by Equation~\ref{equation:alignment}. 
The execution in the corresponding database guarantees the effectiveness of synthesized data $(y',\,\lambda_{i})$.
To synthesize naturalistic training data that reflects LLM prediction behavior, we apply the perturbation operators in Equation~\ref{equation:perturbation} to correctly predicted SQLs $\hat{y}$ to obtain erroneous variants.
To further reflect complex error distributions, we compose multiple operators $\operatorname{\textsc{Rule}}_{\lambda} \circ \operatorname{\textsc{Rule}}_{\lambda'}, \lambda \neq \lambda'$ to produce compound-error samples $(y',\,\{\lambda,\,\lambda'\})$.

\noindent \textbf{(ii) LLM-assisted error injections.}
For incorrect predictions that cannot be attributed to a single perturbation rule, we feed both $\hat{y}$ and the ground-truth $y$ into an assistant model \textsc{AsiLLM} to produce the error type annotation and a refined SQL $\tilde{y}$ based on $\hat{y}$:
\begin{equation}
\label{equation:llm-assist}
    (\tilde{y},\,\{\lambda_i\}) = \operatorname{\textsc{AsiLLM}}(q, \mathcal{S},\,\hat{y}, y),\, \operatorname{\textsc{Exec}}(\tilde{y},\,\mathcal{D}) = \operatorname{\textsc{Exec}}(y,\,\mathcal{D}).
\end{equation}
We accept the annotated $\{\lambda_i\}$ as labels for $\hat{y}$ only when the refined SQL passes execution verification, indicating that the assistant model has faithfully identified the error semantics.

\stitle{Representation Learning of Error Tokens.}
We collect the erroneous SQLs synthesized as our training data of \textsc{ErrorLLM} via Equations~\ref{equation:perturbation} and~\ref{equation:llm-assist} and unify them under a common notation $\bar{y}$. 
We then construct a training set $\{(\mathbf{x}_j^{(\textsc{Det})},\,\ell_j)\}_{j=1}^{M}$ such that:
\begin{equation}
\label{equation:concanate}
    \mathbf{x}_j^{(\textsc{Det})} = [\mathcal{I}_{\Lambda} \parallel q \parallel \mathcal{G} \parallel \bar{y}_{j} \parallel \tau_{\bar{y}_{j}} \parallel \operatorname{\textsc{Exec}}(\bar{y}_{j},\,\mathcal{D})],\,\ell_j \subset \Lambda \cup \{\lambda_\varnothing\}
\end{equation}
where $\mathcal{I}_{\Lambda}$ contains the error type definitions mapping each $\lambda \in \Lambda$ to its semantic description, as detailed in Table~\ref{table:taxonomy}; $\mathbf{x}_j^{(\textsc{Det})}$ is the linearized input sequence and $\ell_j$ is the corresponding label set.
The target output is a set of error tokens $o_j = \{[\textsc{Err}]_{i} \mid \lambda_i \in \ell_j\}$, or $\{[\textsc{Err}]_{\varnothing}\}$ if $\ell_j = \{\lambda_\varnothing\}$.
Since error detection is inherently a set prediction problem in which the output order carries no semantic meaning, we adopt a permutation-invariant training objective:
\begin{equation}
    \mathcal{L} = -\sum_{j=1}^{M} \frac{1}{|o_j|} \sum_{k=1}^{|o_j|} \log\, p_{\theta}\!\left(o_j^{(k)} \mid \mathbf{x}_j,\, o_j^{(<k)}\right),
\end{equation}
where $o_j^{(1)}, \dots, o_j^{(|o_j|)}$  denotes an arbitrary serialization of the label set, and 
$\frac{1}{|o_j|}$ normalizes the loss across label sets of varying lengths.
We fine-tune the model $\operatorname{\textsc{ErrorLLM}}_{\theta}$ with parameter $\theta$ using LoRA adapters~\cite{hu2022lora}, while keeping the error token embedding matrix $\mathbf{E}_{\mathcal{W}' \setminus \mathcal{W}}$ trainable and the original embedding matrix $\mathbf{E}_{\mathcal{W}}$ frozen:
\begin{equation}
    \theta' = \theta \cup \Delta\theta_{\textsc{LoRA}} \cup \mathbf{E}_{\mathcal{W}' \setminus \mathcal{W}}.
\end{equation}
This strategy preserves the model's pre-trained linguistic and SQL generation capabilities while allowing the error tokens to learn discriminative representations within the shared semantic space.

\subsection{Detecting SQL Errors via \textsc{ErrorLLM}}
\label{section:detection}
Given the QSS $\mathcal{G}$, the AST $\tau_{\hat{y}}$, and the corresponding execution feedback $\operatorname{\textsc{Exec}}(\hat{y},\,\mathcal{D})$, the goal of error detection is to produce a set of detected error types $\{\hat{\lambda}\} \subseteq \Lambda \cup \{\lambda_\varnothing\}$, 
where $\{\hat{\lambda}\} = \{\lambda_\varnothing\}$ indicates that no error is detected.
The detection process of our proposed method consists of two hierarchical stages introduced below.

\stitle{Static Superficial Detection.}
The first stage applies static symbolic rules to perform structural surface validation between the AST and QSS, which identifies obvious structural mismatches and captures explicit execution errors from the feedback~\cite{li2025deepeye}.
Following the perturbation operators defined in Section~\ref{section:modeling}, we construct the corresponding detection rules by inversion: each $\operatorname{\textsc{Rule}}_{\lambda_i}$ injects a specific structural violation into a correct AST, and inverting it yields a deterministic predicate that checks for the presence of the same violation in a predicted AST $\tau_{\hat{y}}$.
We term each such inverted predicate a \textbf{detection rule} and construct the detection rule set $\mathcal{R}=\{\overleftarrow{\operatorname{\textsc{Rule}}}_{\lambda} \mid \lambda \in \Lambda\}$, where each $\overleftarrow{\operatorname{\textsc{Rule}}}_{\lambda}$ checks whether an AST node violates a specific constraint implied by $\mathcal{G}$ and/or $\operatorname{\textsc{Exec}}(\hat{y},\,\mathcal{D})$.
Formally, we traverse all nodes $v\in\tau_{\hat{y}}$ by applying each detection rule and collect the triggered error types:
\begin{equation}
    \{\hat{\lambda}\}_{\textsc{Rule}}=\left\{ \lambda \mid \forall v \not\models \overleftarrow{\operatorname{\textsc{Rule}}}_{\lambda},\,v\in\tau_{\hat{y}},\,\overleftarrow{\operatorname{\textsc{Rule}}}_{\lambda}\in\mathcal{R}\right\},
\end{equation}
where $v \not\models \overleftarrow{\operatorname{\textsc{Rule}}}_{\lambda}$ denotes that node $v$ violates rule $\overleftarrow{\operatorname{\textsc{Rule}}}_{\lambda}$
The rule-based detection result $\{\hat{\lambda}\}_{\textsc{Rule}}$ serves both as an explicit detection output and as auxiliary input to the subsequent LLM-based detection stage.
If no error is explicitly detected, we have $\{\hat{\lambda}\}_{\textsc{Rule}} =\emptyset$.

\stitle{LLM-based Semantic Detection.}
To further detect implicit semantic errors that cannot be captured by static rules, we introduce LLM-based detection to capture semantic inconsistencies that lie beyond execution failures and surface mismatches.
Following Section~\ref{section:modeling}, the fine-tuned $\operatorname{\textsc{ErrorLLM}}_{\theta'}$ predicts all error types present in $\tau_{\hat{y}}$ conditioned on the structural input $\mathbf{x}^{(\textsc{Det})}$ constructed via Equation~\ref{equation:concanate}, augmented with the static detection result $\{\hat{\lambda}\}_{\textsc{Rule}}$.
The model generates a sequence of error tokens as follows:
 \begin{equation}
    o = \operatorname{\textsc{ErrorLLM}}_{\theta'}\!\left(\mathbf{x}^{(\textsc{Det})}\right) = [\textsc{Err}]_{i_1} \cdots [\textsc{Err}]_{i_l},
\end{equation}
where each $[\textsc{Err}]_{i}$ is defined using Equation~\ref{equation:error-assign} that maps to an error with $l \ge 1$, or $o = [\textsc{Err}]_{\varnothing}$ if no error is detected.

During inference, we apply constrained decoding to ensure that the output consists exclusively of valid error tokens.
We restrict the output distribution to the extended error token set:
\begin{equation}
    \hat{p}_{\theta'}\!\left(w \mid \mathbf{x}, o^{(<k)}\right) =
    \begin{cases}
    \dfrac{p_{\theta'}\!\left(w \mid \mathbf{x}, o^{(<k)}\right)}{\sum_{s \in \mathcal{W}' \setminus \mathcal{W}} p_{\theta'}\!\left(s \mid \mathbf{x}, o^{(<k)}\right)}, & w \in \mathcal{W}' \setminus \mathcal{W} \\
    0, & w \in \mathcal{W}
    \end{cases},
\end{equation}
where $p_{\theta'}$ denotes the raw output distribution of $\operatorname{\textsc{ErrorLLM}}_{\theta'}$ 
and $\hat{p}_{\theta'}$ denotes the constrained distribution restricted to the error token vocabulary 
$\mathcal{W}' \setminus \mathcal{W}$, ensuring that the model produces only well-formed error token sequences without extraneous natural language output.
The detected error types from the LLM can be presented as $\{\hat{\lambda}\}_{\textsc{LLM}} = \{\lambda_i \mid [\textsc{Err}]_{i} \in o\}$.
The final detection output is given by $\{\hat{\lambda}\} = \{\hat{\lambda}\}_{\textsc{Rule}} \cup \{\hat{\lambda}\}_{\textsc{LLM}}$, which serves as the foundation for the subsequent error-guided text-to-SQL refinement stage.
The pseudo-code of the SQL error detection is provided in Algorithm~\ref{algorithm:detection}.
\begin{algorithm}[!t]
\caption{SQL Error Detection}
\label{algorithm:detection}
\begin{algorithmic}[1]
\Require Predicted SQL $\hat{y}$, AST $\tau_{\hat{y}}$, QSS $\mathcal{G}$,
         execution feedback $\operatorname{\textsc{Exec}}(\hat{y},\mathcal{D})$,
         structural input $\mathbf{x}^{(\textsc{Det})}$,
         detection rule set $\mathcal{R} = \{\overleftarrow{\operatorname{\textsc{Rule}}}_\lambda \mid \lambda \in \Lambda\}$,
         fine-tuned $\operatorname{\textsc{ErrorLLM}}_{\theta'}$
\Ensure Detected error type set $\{\hat{\lambda}\} \subseteq \Lambda \cup \{\lambda_\varnothing\}$

\Statex \textcolor{gray}{\textit{\% Static Superficial Detection}}
\State $\{\hat{\lambda}\}_{\textsc{Rule}} \gets \emptyset$
\For{each $v \in \mathcal{V}_{\tau_{\hat{y}}}$}
    \For{each $\overleftarrow{\operatorname{\textsc{Rule}}}_\lambda \in \mathcal{R},$ where $\lambda \in \Lambda$}
        \If{$v \not\models \overleftarrow{\operatorname{\textsc{Rule}}}_\lambda$}
            \State $\{\hat{\lambda}\}_{\textsc{Rule}} \gets \{\hat{\lambda}\}_{\textsc{Rule}} \cup \{\lambda\}$
        \EndIf
    \EndFor
\EndFor
\Statex \textcolor{gray}{\textit{\% LLM-based Semantic Detection}}
\State Augment input with rule results: $\mathbf{x}^{(\textsc{Det})} \gets [\mathbf{x}^{(\textsc{Det})} \parallel \{\hat{\lambda}\}_{\textsc{Rule}}]$
\State Generate error token sequence with constrained decoding:
\[
    \hat{p}_{\theta'}\!\left(w \mid \mathbf{x}, o^{(<k)}\right) =
    \begin{cases}
    \dfrac{p_{\theta'}\!\left(w \mid \mathbf{x}, o^{(<k)}\right)}{\sum_{s \in \mathcal{W}' \setminus \mathcal{W}} p_{\theta'}\!\left(s \mid \mathbf{x}, o^{(<k)}\right)}, & w \in \mathcal{W}' \setminus \mathcal{W} \\
    0, & w \in \mathcal{W}
    \end{cases}
\]
\State $o \gets \operatorname{\textsc{ErrorLLM}}_{\theta'}\!\left(\mathbf{x}^{(\textsc{Det})}\right)$
\State $\{\hat{\lambda}\}_{\textsc{LLM}} \gets \{\lambda_i \mid [\textsc{Err}]_i \in o\}$

\Statex \textcolor{gray}{\textit{\% Result Aggregation}}
\State $\{\hat{\lambda}\} \gets \{\hat{\lambda}\}_{\textsc{Rule}} \cup \{\hat{\lambda}\}_{\textsc{LLM}}$
\If{$\{\hat{\lambda}\} = \emptyset$}
    \State $\{\hat{\lambda}\} \gets \{\lambda_\varnothing\}$
\EndIf
\State \Return $\{\hat{\lambda}\}$
\end{algorithmic}
\end{algorithm}

\subsection{Error-guided Text-to-SQL Refinement}
Given the detected error type set $\{\hat{\lambda}\}$ from error detection, the goal of text-to-SQL refinement is to fix all identified errors in $\hat{y}$ and produce a corrected SQL.
Directly prompting an LLM with raw errors often leads to superficial fixes, as the model lacks fine-grained context about where and why each error occurs.
We propose an error-guided refinement pipeline that decomposes the task into three stages through a dual-LLM architecture introduced below.

\stitle{Error Localization and Analysis.}
To bridge the detected error types $\{\hat{\lambda}\}$ with actionable error-guided refinement, we employ a localization model $\operatorname{{\textsc{LocLLM}}}$ that performs fine-grained error analysis.
Each error type $\lambda \in \Lambda$ is associated with a pre-defined guideline template $g_{\lambda}[\cdot]$ with fillable slots that structure the error analysis (e.g., current tables, missing table, error type for a ``table missing'' error).
The input is constructed by concatenating the structural input $\mathbf{x}^{(\textsc{Det})}$ with $\{\hat{\lambda}\}$ and the corresponding guideline templates:
\begin{equation}
    \mathbf{x}^{(\textsc{Loc})} = [\mathbf{x}^{(\textsc{Det})} \parallel \{\hat{\lambda}\} \parallel \{g_{\lambda_m}[\cdot]\}_{\lambda_m \in \{\hat{\lambda}\}}].
\end{equation}
Then $\operatorname{{\textsc{LocLLM}}}$ jointly analyzes all detected errors and produces:
\begin{equation}
    \{(\mathcal{V}_{\tau_{\hat{y}}}^{(\lambda_m)},\, \mathcal{S}^{(\lambda_m)},\, g_{\lambda_m}[\ast])\}_{\lambda_m \in \{\hat{\lambda}\}} = \operatorname{{\textsc{LocLLM}}}\!\left(\mathbf{x}^{(\textsc{Loc})}\right),
\end{equation}
where $\mathcal{V}_{\tau_{\hat{y}}}^{(\lambda_m)} \subset \mathcal{V}_{\tau_{\hat{y}}}$ denotes the localized error nodes in the AST; $\mathcal{S}^{(\lambda_m)} \subset \mathcal{S}$ denotes the involved schema elements; and $g_{\lambda_m}[\ast]$ is the completed guideline with all slots filled, for each $\lambda_m \in \{\hat{\lambda}\}$.

\stitle{Error-guided SQL Refinement.}
Given the localization results, we combine the refinement context by extracting relevant content for each detected error $\lambda_m \in \{\hat{\lambda}\}$:
\begin{equation}
    (\tau_{\hat{y}}^{(\lambda_m)},\, \mathcal{G}^{(\lambda_m)},\, {\epsilon}^{(\lambda_m)}) = \operatorname{\textsc{Ext}}(\tau_{\hat{y}},\, \mathcal{G},\, \mathcal{V}_{\tau_{\hat{y}}}^{(\lambda_m)},\, \mathcal{S}^{(\lambda_m)},\, \lambda_m),
\end{equation}
where $\tau_{\hat{y}}^{(\lambda_m)} \subseteq \tau_{\hat{y}}$ is the minimal AST subtree enclosing the error, $\mathcal{G}^{(\lambda_m)} \subseteq \mathcal{G}$ is the relevant schema structure, and $\epsilon^{(\lambda_m)}$ is a set of retrieved few-shot examples demonstrating the SQL refinement process.
The scope of extraction is determined by $\lambda_m$.

\begin{algorithm}[!t]
\caption{Text-to-SQL Refinement}
\label{algorithm:refinement}
\begin{algorithmic}[1]
\Require Predicted SQL $\hat{y}$, AST $\tau_{\hat{y}}$, QSS $\mathcal{G}$,
structural input $\mathbf{x}^{(\textsc{Det})}$,
         detected error type set $\{\hat{\lambda}\}$ from Algorithm~\ref{algorithm:detection},
         guideline templates $\{g_{\lambda}[\cdot]\}_{\lambda \in \Lambda}$
\Ensure Refined SQL $\tilde{y}$
\If{$\{\hat{\lambda}\} = \{\lambda_\varnothing\}$}
    \State \Return $\hat{y}$
\EndIf
\Statex \textcolor{gray}{\textit{\% Error Localization and Analysis}}
\State $\mathbf{x}^{(\textsc{Loc})} \gets [\mathbf{x}^{(\textsc{Det})} \parallel \{\hat{\lambda}\} \parallel \{g_{\lambda_m}[\cdot]\}_{\lambda_m \in \{\hat{\lambda}\}}]$
\State $\{(\mathcal{V}_{\tau_{\hat{y}}}^{(\lambda_m)},\, \mathcal{S}^{(\lambda_m)},\, g_{\lambda_m}[\ast])\}_{\lambda_m \in \{\hat{\lambda}\}} \gets \operatorname{\textsc{LocLLM}}\!\left(\mathbf{x}^{(\textsc{Loc})}\right)$
\Statex \textcolor{gray}{\textit{\% Error-guided SQL Refinement}}
\State $\mathbf{x}^{(\textsc{Ref})} \gets [\,]$
\For{each $\lambda_m \in \{\hat{\lambda}\}$}
    \State $(\tau_{\hat{y}}^{(\lambda_m)},\, \mathcal{G}^{(\lambda_m)},\, \epsilon^{(\lambda_m)}) \gets \operatorname{\textsc{Ext}}(\tau_{\hat{y}},\, \mathcal{G},\, \mathcal{V}_{\tau_{\hat{y}}}^{(\lambda_m)},\, \mathcal{S}^{(\lambda_m)},\, \lambda_m)$
    \State $\mathbf{x}^{(\textsc{Ref})}.\text{append}\!\left((\tau_{\hat{y}}^{(\lambda_m)},\, \mathcal{G}^{(\lambda_m)},\, g_{\lambda_m}[\ast],\, \epsilon^{(\lambda_m)})\right)$
\EndFor
\State $\mathbf{x}^{(\textsc{Ref})} \gets \textsc{Sort}(\mathbf{x}^{(\textsc{Ref})},\, \phi)$ 
\State $\tilde{y} \gets \operatorname{\textsc{RefLLM}}(\hat{y},\, \mathbf{x}^{(\textsc{Ref})})$
\State \Return $\tilde{y}$
\end{algorithmic}
\end{algorithm}
We then order all error contexts by a priority function $\phi: \Lambda \rightarrow \mathbb{N}$ that reflects inter-error dependencies, where structural errors (e.g., ``table missing'') receive the highest priority as they may induce cascading downstream errors.
The priority-ordered context is:
\begin{equation}
    \mathbf{x}^{(\textsc{Ref})} = \left[ (\tau_{\hat{y}}^{(\lambda_m)},\, \mathcal{G}^{(\lambda_m)},\, g_{\lambda_m}[\ast],\, \epsilon^{(\lambda_m)}) \right]_{\lambda_m \in \{\hat{\lambda}\}},
\end{equation}
where the order follows $\phi(\lambda_1) \leq \cdots \leq \phi(\lambda_{|\{\hat{\lambda}\}|})$
Finally, the refinement model $\operatorname{\textsc{RefLLM}}$ receives $\mathbf{x}^{(\textsc{Ref})}$ togeter with the original SQL $\hat{y}$ and produces the refined SQL $\tilde{y}$ in a single forward pass:
\begin{equation}
    \tilde{y} = \operatorname{\textsc{RefLLM}}\!\left(\mathbf{x}^{(\textsc{Ref})},\,\hat{y}\right).
\end{equation}
Presenting all errors simultaneously in priority order allows the LLM to reason about complex inter-error dependencies in a single pass, avoiding the hallucination accumulation risks inherent in iterative correction without guidance.
The pseudo-code of the text-to-SQL refinement workflow is provided in Algorithm~\ref{algorithm:refinement}.

\section{Experiments}

To comprehensively evaluate \textsc{ErrorLLM}, we design experiments around four research questions.
\textbf{RQ1}: How does \textsc{ErrorLLM} perform in end-to-end text-to-SQL refinement compared with existing methods?
\textbf{RQ2}: How effective are the SQL error detection task and the text-to-SQL refinement task individually and jointly?
\textbf{RQ3}: How accurate is \textsc{ErrorLLM}'s fine-grained error type prediction at the per-category level?
\textbf{RQ4}: What is the contribution of each key component to the overall performance?

\begin{table*}[!t]
\centering
\caption{Results of text-to-SQL execution accuracy (EX) (\%) based on GPT-4o (above the double divider) and OpenSearch-SQL (below the double divider) generated SQLs on BIRD~\cite{li2023bird} and \textsc{Spider}~\cite{yu2018spider} development set.
Each baseline represents a specific text-to-SQL refinement method applied to the GPT-4o/OpenSearch-SQL outputs.
Bold and underlined values indicate the best and second-best results, respectively.
Values in parentheses denote performance improvements (green) and degradations (red) in percentage relative to the baseline.
Categories Simp., Mode., and Chal. denote Simple, Moderate, and Challenging difficulty levels within the BIRD dataset.
The refinement backbone are indicated within parentheses after \textsc{ErrorLLM}
$^{\diamondsuit}$ The self-correction module is vanilla implementation~\cite{huang2023large}.
$^{\clubsuit}$ The self-correction in DIN-SQL~\cite{pourreza2023dinsql} is set with the gentle prompt.}
\label{table:main}
\vspace{-3mm}
\setlength{\tabcolsep}{3pt}
\resizebox{1.0\textwidth}{!}{
    \begin{tabular}{lcccccccccc}
        \toprule
        \multirow{2}{*}{\textbf{Methods}} & \multicolumn{4}{c}{\textbf{\textsc{BIRD}}} & \multicolumn{5}{c}{\textbf{\textsc{Spider}}} \\
        \cmidrule(lr){2-5} \cmidrule(lr){6-10}
        & \textbf{Simp.} & \textbf{Mode.} & \textbf{Chal.} & \textbf{Overall} & \textbf{Easy} & \textbf{Medium} & \textbf{Hard} & \textbf{Extra} & \textbf{Overall} \\
        \midrule
        GPT-4o~\cite{qu2025share} & 63.35 & 44.18 & 45.52 & 55.87 & 91.53 & 72.42 & 76.44 & 58.43 & 75.44 \\
        
        \midrule
        +$\text{Self-correction}^{\diamondsuit}$~\cite{huang2023large} & 62.92 (\negnum{-0.68\%}) & 43.32 (\negnum{-1.95\%}) & 42.07 (\negnum{-7.58\%}) & 55.02 (\negnum{-1.52\%}) & 90.73 (\negnum{-0.87\%}) & 71.75 (\negnum{-0.93\%}) & 75.29 (\negnum{-1.50\%}) & 56.02 (\negnum{-4.12\%}) & 74.47 (\negnum{-1.29\%}) \\
        
        +Self-debugging~\cite{chen2024teach} & 64.00 (\pos{1.03\%}) & 45.04 (\pos{1.95\%}) & 46.90 (\pos{3.03\%}) & 56.65 (\pos{1.40\%}) & 93.55 (\pos{2.21\%}) & 74.22 (\pos{2.49\%}) & 77.01 (\pos{0.75\%}) & 58.43 ($\pm$0.00\%) & 76.79 (\pos{1.79\%}) \\
        
        +Self-consistency~\cite{dong2023c3} & 65.73 (\pos{3.76\%}) & 49.14 (\pos{11.23\%}) & 44.14 (\negnum{-3.03\%}) & 58.67 (\pos{5.01\%}) & 92.74 (\pos{1.32\%}) & 73.09 (\pos{0.93\%}) & 77.59 (\pos{1.50\%}) & 59.04 (\pos{1.04\%}) & 76.31 (\pos{1.15\%}) \\
        
        +Cross-consistency~\cite{li2025pet} & 65.84 (\pos{3.93\%}) & 49.57 (\pos{12.20\%}) & 45.52 ($\pm$0.00\%) & 59.00 (\pos{5.60\%}) & 93.15 (\pos{1.77\%}) & 72.87 (\pos{0.62\%}) & 78.16 (\pos{2.25\%}) & 60.24 (\pos{3.10\%}) & 76.60 (\pos{1.54\%}) \\
        
        +Multiple-prompt~\cite{lee2025mcs} & 66.38 (\pos{4.78\%}) & 48.06 (\pos{8.78\%}) & 44.83 (\negnum{-1.52\%}) & 58.80 (\pos{5.24\%}) & 91.94 (\pos{0.45\%}) & 73.32 (\pos{1.24\%}) & 78.74 (\pos{3.01\%}) & 57.83 (\negnum{-1.03\%}) & 76.21 (\pos{1.02\%}) \\

        +$\text{Self-correction}^{\clubsuit}$~\cite{pourreza2023dinsql} & 65.73 (\pos{3.76\%}) & 46.98 (\pos{6.34\%}) & 46.21 (\pos{1.52\%}) & 58.21 (\pos{4.19\%}) & 93.15 (\pos{1.77\%}) & 75.11 (\pos{3.71\%}) & 78.16 (\pos{2.25\%}) & 59.64 (\pos{2.07\%}) & 77.47 (\pos{2.69\%}) \\
        
        \midrule
        +Refiner~\cite{wang2024macsql} & 64.22 (\pos{1.37\%}) & 45.91 (\pos{3.92\%}) & 47.59 (\pos{4.55\%}) & 57.11 (\pos{2.22\%}) & 93.15 (\pos{1.77\%}) & 78.70 (\pos{8.67\%}) & 77.01 (\pos{0.75\%}) & 58.43 ($\pm$0.00\%) & 78.63 (\pos{4.23\%})\\
        
        +SQLFixAgent~\cite{cen2025sqlfixagent} & 65.62 (\pos{3.58\%}) & 48.28 (\pos{9.28\%}) & \underline{48.28} (\pos{6.06\%}) & 58.74 (\pos{5.14\%}) & 93.55 (\pos{2.21\%}) & 90.36 (\pos{24.77\%}) & 77.59 (\pos{1.50\%}) & \underline{69.28} (\pos{18.57\%}) & 85.59 (\pos{13.45\%}) \\
        
        +MAGIC~\cite{askari2025magic} & 67.57 (\pos{6.66\%}) & 49.57 (\pos{12.20\%}) & 46.21 (\pos{1.52\%}) & 60.10 (\pos{7.57\%}) & -- & -- & -- & -- & 85.69 (\pos{13.59\%}) \\
        
        +SHARE~\cite{qu2025share} & 70.81 (\pos{11.78\%}) & 56.25 (\pos{27.32\%}) & 46.90 (\pos{3.03\%}) & 64.14 (\pos{14.80\%}) & 93.95 (\pos{2.64\%}) & 90.13 (\pos{24.45\%}) & \underline{78.16} (\pos{2.25\%}) & \textbf{70.48} (\pos{20.62\%}) & 85.88 (\pos{13.84\%}) \\
        
        \midrule
        \textbf{+\textsc{ErrorLLM} (DeepSeek-V3)} & \underline{71.08} (\pos{12.20\%}) & \underline{56.90} (\pos{28.79\%}) & 47.59 (\pos{4.55\%}) & \underline{64.99} (\pos{16.32\%}) & \underline{94.76} (\pos{3.53\%}) & \textbf{91.26} (\pos{26.01\%}) & \underline{78.16} (\pos{2.25\%}) & \underline{69.28} (\pos{18.57\%}) & \underline{86.36} (\pos{14.48\%})\\
        \textbf{+\textsc{ErrorLLM} (GPT-4o)} & \textbf{73.08} (\pos{15.36\%}) & \textbf{57.97} (\pos{31.21\%}) & \textbf{48.97} (\pos{7.58\%}) & \textbf{66.23} (\pos{18.54\%}) & \textbf{95.56} (\pos{4.40\%}) & \underline{91.03} (\pos{25.70\%}) & \textbf{79.89} (\pos{4.51\%}) & \textbf{70.48} (\pos{20.62\%}) & \textbf{86.94} (\pos{15.24\%})\\
        
        \midrule
        \midrule
        OpenSearch-SQL~\cite{xie2025open} & 75.46 & 64.66 & \textbf{66.90} & 71.38 & 92.74 & 88.57 & \underline{79.31} & \textbf{69.88} & 85.01 \\
        
        \midrule
        +$\text{Self-correction}^{\clubsuit}$~\cite{pourreza2023dinsql} & 67.35 (\negnum{-10.75\%}) & 54.31 (\negnum{-16.01\%}) & 44.83 (\negnum{-32.99\%}) & 61.28 (\negnum{-14.15\%}) & -- & -- & -- & -- & -- \\
        
        +SQLFixAgent~\cite{cen2025sqlfixagent} & 74.05 (\negnum{-1.87\%}) & 60.34 (\negnum{-6.68\%}) & 62.07 (\negnum{-7.22\%}) & \underline{68.77} (\negnum{-3.66\%}) & 93.55 (\pos{0.87\%}) & 89.01 (\pos{0.50\%}) & 78.16 (\negnum{-1.45\%}) & \underline{67.47} (\negnum{-3.45\%}) & 84.82 (\negnum{-0.22\%}) \\
        
        +SHARE~\cite{qu2025share} & 73.19 (\negnum{-3.01\%}) & 57.76 (\negnum{-10.67\%}) & 47.59 (\negnum{-28.86\%}) & 66.10 (\negnum{-7.40\%}) & -- & -- & -- & -- & -- \\
        
        \midrule
        \textbf{+\textsc{ErrorLLM} (DeepSeek-V3)} & \underline{76.43} (\pos{1.29\%}) & \underline{65.09} (\pos{0.67\%}) & 64.83 (\negnum{-3.09\%}) & \underline{71.90} (\pos{0.73\%}) & \underline{95.16} (\pos{2.61\%}) & \textbf{91.48} (\pos{3.29\%}) & 78.73 (\negnum{-0.73\%}) & \textbf{69.88} ($\pm$0.00\%) & \underline{86.75} (\pos{2.05\%})\\
        \textbf{+\textsc{ErrorLLM} (GPT-4o)} & \textbf{76.65} (\pos{1.58\%}) & \textbf{65.52} (\pos{1.33\%}) & \underline{66.21} (\negnum{-1.03\%}) & \textbf{72.29} (\pos{1.27\%}) & \textbf{95.97} (\pos{3.48\%}) & \underline{91.26} (\pos{3.04\%}) & \textbf{79.89} (\pos{0.73\%}) & \textbf{69.88} ($\pm$0.00\%) & \textbf{87.04} (\pos{2.39\%})\\
        
        \bottomrule
    \end{tabular}
}
\end{table*}

\subsection{Experimental Setups}
\label{section:setups}
\stitle{Datasets.}
First, we conduct the text-to-SQL refinement experiments on two widely-used text-to-SQL benchmarks.
\textbf{(i) \textsc{BIRD}}~\cite{li2023bird} is a challenging text-to-SQL benchmark, and we evaluate on the official development set (1,534 samples).
\textbf{(ii) \textsc{Spider}}~\cite{yu2018spider} contains standard databases for text-to-SQL evaluation, and we also evaluate on the official development set (1,034 samples).
To further examine generalization capability, we additionally evaluate on three
\textbf{(iii) \textsc{Spider variants}}: \textsc{Spider-Realistic}~\cite{deng2021structure} (508 samples), \textsc{Spider-Syn}~\cite{gan2021towards} (1034 samples), and \textsc{Spider-DK}~\cite{gan2021exploring} (535 samples), which introduce perturbations in question phrasing, synonym substitution, and domain knowledge requirements, respectively.
Secondly, we conduct SQL error detection experiments on GPT-4o generated SQLs and a dedicated SQL error detection benchmark:
\textbf{(iv) \textsc{NL2SQL-Bugs}}~\cite{liu2025nl2sqlbug} contains expert-annotated instances that detail the SQL errors for semantically incorrect user questions.
The experiments of \textsc{NL2SQL-Bugs} are also based on \textsc{BIRD} development set (2,018 samples).
Since the pre-defined error types in \textsc{NL2SQL-Bugs} are largely different, we map our error taxonomy $\Lambda$ to their categories for evaluation.
Unless otherwise specified, RQ1 is evaluated on all three benchmarks (i)--(iii) across both backbones on (i), and across GPT-4o on (ii) and (iii); RQ2, RQ4, and the extended analyses in the Appendix are conducted on the \textsc{BIRD} development set; and RQ3 is evaluated on the dedicated eroor detection benchmark \textsc{NL2SQL-Bugs}.

\stitle{Evaluation Metrics.}
\label{section:metrics}
We adopt the following metrics across the proposed research questions.
\textbf{(i) Execution accuracy (EX)}~\cite{yu2018spider} measures whether the predicted SQL returns the same result as the ground-truth SQL when executed against the database, serving as the primary metric for end-to-end text-to-SQL refinement evaluation (RQ1, RQ4).
\textbf{(ii) Detection accuracy (D-Accuracy)} measures the correctness of binary error detection for each SQL sample.
\textbf{(iii) Detection F1 Score (D-F1)} as the harmonic mean of \textbf{Precision} and \textbf{Recall}, evaluates the quality of SQL error detection (RQ2, RQ4).
\textbf{(iv) Fixed Rate (FR)} measures the proportion of the truly incorrect SQLs that were detected (true positives, TP) that are successfully fixed after refinement (RQ2):
\begin{equation}
    \text{FR} = \frac{\left|\left\{\hat{y} \in \text{TP} \mid \operatorname{\textsc{Exec}}(\tilde{y}, \mathcal{D}) = \operatorname{\textsc{Exec}}(y, \mathcal{D})\right\}\right|}{|\text{TP}|}.
\end{equation}
\textbf{(v) Corruption Rate (CR)} measures the proportion of falsely detected correct SQLs (false positives, FP) whose execution results change after refinement (RQ2):
\begin{equation}
    \text{CR} = \frac{\left|\left\{\hat{y} \in \text{FP} \mid \operatorname{\textsc{Exec}}(\tilde{y}, \mathcal{D}) \neq \operatorname{\textsc{Exec}}(\hat{y}, \mathcal{D})\right\}\right|}{|\text{FP}|}.
\end{equation}
\textbf{(vi) Type-specific Accuracy (TSA)}~\cite{liu2025nl2sqlbug} measures per-category detection accuracy for fine-grained error type evaluation (RQ3).

\stitle{Baselines.}
We compare \textsc{ErrorLLM} with the following categories of baselines for our main experiment: \textbf{(i) Self-correction}: The vanilla self-correction implementation~\cite{huang2023large} and the self-correction module in DIN-SQL~\cite{pourreza2023dinsql} with gentle prompt style, which leverage LLMs' internal reasoning.
\textbf{(ii) Self-debugging}: Self-debugging~\cite{chen2024teach}, which iteratively refines SQLs using execution feedback.
\textbf{(iii) Self-consistency}: The vanilla self-consistency module~\cite{dong2023c3}, the cross-consistency module from PET-SQL~\cite{li2025pet}, and the multiple-prompt module from MCS-SQL~\cite{lee2025mcs}, which select or vote among multiple SQL candidates.
\textbf{(iv) Dedicated Refinement}: The Refiner module from MAC-SQL~\cite{wang2024macsql}, SQLFixAgent~\cite{cen2025sqlfixagent}, MAGIC~\cite{askari2025magic}, and SHARE~\cite{qu2025share}, which design dedicated pipelines for SQL refinement.
To ensure the consistency, the baselines are implemented by GPT-4o.
For the fine-grained error type evaluation (RQ3), we additionally compare with proprietary LLMs including GPT-4o, GPT-4o-mini, Gemini-2.0-Flash, and DeepSeek-V3, evaluated under the \textsc{NL2SQL-Bugs}~\cite{liu2025nl2sqlbug} benchmark setting (with error types mapping introduced below).

\stitle{Implementations.}
We use GPT-4o~\cite{qu2025share} and OpenSearch-SQL~\cite{xie2025open} as the backbone methods.
The error detection model $\textsc{ErrorLLM}$ is trained from CodeS-7B~\cite{li2024codes} using LoRA~\cite{hu2022lora} adapters.
The error token vocabulary is initialized with $N = 32$ reserved tokens, where the current taxonomy covers $|\Lambda| = 12$ error types, as detailed in Table~\ref{table:taxonomy}.
For the localization model $\textsc{LocLLM}$ and the refinement model $\textsc{RefLLM}$, we use GPT-4o as the default backbone.
Training data is synthesized from the \textsc{BIRD} training set.
All experiments are conducted using 4$\times$NVIDIA A100-SXM4-80GB GPUs.
Since the pre-defined error types in \textsc{NL2SQL-Bugs} are different, the mapping details are in Appendix~\ref{appendix:taxonomy} and in our repository.

\subsection{RQ1: End-to-End Text-to-SQL Refinement}
\label{section:rq1}

Table~\ref{table:main} presents the end-to-end execution accuracy of the text-to-SQL refinement task on \textsc{BIRD} and \textsc{Spider} development sets for SQLs generated by two backbones: GPT-4o~\cite{qu2025share} and OpenSearch-SQL~\cite{xie2025open}. 
We highlight the following observations.

\begin{figure}[!t]
    \centering
    \begin{tikzpicture}[scale=0.8]
    \begin{axis}[
        height=5.5cm, 
        width=1.15\linewidth,
        ybar = .17mm, 
        enlarge y limits={upper=0.2}, 
        enlarge x limits=0.20, 
        ymin = 50, 
        ymax = 100,
        ytick={50, 60, 70, 80, 90, 100},
        axis line style=ultra thick,
        xticklabel={font=\small},
        yticklabel style={font=\small},
        ylabel={Execution Accuracy (\%)},
        ylabel style={yshift=-0.3ex},
        symbolic x coords={Spider, Realistic, Syn, DK},
        xtick=data,
        xticklabels={\textsc{Spider}, \textsc{Realistic}, \textsc{Syn}, \textsc{Dk}},
        legend image code/.code={%
            \draw[#1, draw=black] (0cm,-0.1cm) rectangle (0.28cm,0.08cm);
        },
        bar width=12pt,
        xmajorgrids=true,
        ymajorgrids=true,
        grid style=dashed,
        legend pos=north east,
        legend style={
            font=\small,
            at={(0.5, 0.98)},
            legend columns=-1,
            anchor=north
            }
        ]

    \addplot [
        line width= .3mm, 
        fill=colOne,
        postaction={
            pattern=crosshatch dots, pattern color=white
        }
    ]
    coordinates {(Spider,75.44) (Realistic,67.79) (Syn,73.43) (DK,63.36)};

    \addplot[
        line width= .3mm, 
        fill=colTwo,
        postaction={
            pattern=north east lines, pattern color=white
        }
    ]
    coordinates {(Spider,85.59) (Realistic,72.82) (Syn,75.59) (DK,67.85)};
        
    \addplot [
        line width= .3mm,
        fill=colThree,
        point meta=explicit symbolic,
        every node near coord/.append style={
            font=\scriptsize\bfseries, 
            color=black, 
            yshift=2pt, 
            anchor=south
        },
        postaction={
            pattern=vertical lines, pattern color=white
         }
    ]
    coordinates {(Spider, 86.94) (Realistic,75.63) (Syn,78.15) (DK,69.53)};

    \node[above, font=\footnotesize\bfseries, yshift=1pt, xshift=12pt] at (axis cs:Spider,86.94) {$\uparrow${11.5}};
    \node[above, font=\footnotesize\bfseries, yshift=1pt, xshift=12pt] at (axis cs:Realistic,75.63) {$\uparrow${7.8}};
    \node[above, font=\footnotesize\bfseries, yshift=1pt, xshift=12pt] at (axis cs:Syn,78.15) {$\uparrow${4.7}};
    \node[above, font=\footnotesize\bfseries, yshift=1pt, xshift=12pt] at (axis cs:DK,69.53) {$\uparrow${6.2}};
    
    \legend{GPT-4o, SQLFixAgent, \textbf{\textsc{ErrorLLM}} }
    \end{axis}
    \end{tikzpicture}
    \vspace{-3mm}
    \caption{Results of text-to-SQL execution accuracy (EX) (\%) based on GPT-4o generated SQLs on \textsc{Spider variants}. The values with uparrow indicate improvements of EX.}
    \vspace{-5mm}
\label{figure:variant}
\end{figure}
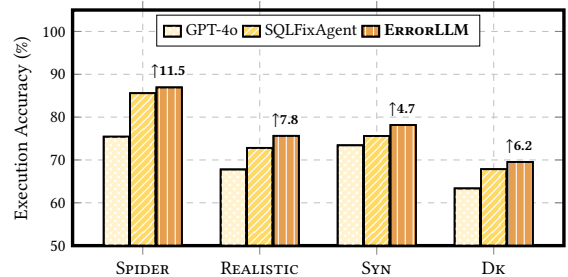
\stitle{Effectiveness across Backbones.}
\textsc{ErrorLLM} achieves the best overall performance on both benchmarks and backbones.
On GPT-4o, \textsc{ErrorLLM} improves execution accuracy by +18.54\% on \textsc{BIRD} (55.87$\to$66.23) and +15.24\% on \textsc{Spider} (75.44$\to$86.94), outperforming the baselines.
The improvements are generally consistent across all difficulty levels, with notable improvements on \textsc{BIRD}-Simple (+15.36\%) and \textsc{BIRD}-Moderate (+31.21\%).
On OpenSearch-SQL, most existing refinement methods degrade performance when applied to a stronger backbone, showing a high corruption rate.
In contrast, \textsc{ErrorLLM} is the only method that generally improves the already-strong OpenSearch-SQL on both datasets (+1.27\% on \textsc{BIRD}, +2.39\% on \textsc{Spider}).
This robustness stems from our explicit error modeling: by detecting whether genuine errors exist before refinement, \textsc{ErrorLLM} avoids the false-positive corrections that other methods struggle with when the initial SQL quality is already high.
Furthermore, we evaluate on three \textsc{Spider variant} benchmarks to examine generalization.
As shown in Figure~\ref{figure:variant}, \textsc{ErrorLLM} consistently outperforms both the GPT-4o backbone and the strongest baseline SQLFixAgent across all variants, achieving execution accuracy value improvements of $\uparrow$7.8, $\uparrow$4.7, and $\uparrow$6.2 on \textsc{Spider-Realistic}, \textsc{Spider-Syn}, and \textsc{Spider-DK}, respectively.
These results demonstrate that our explicit error modeling generalizes well to distribution-shifted settings where question phrasing, synonyms, and domain knowledge requirements differ from the training data.

\stitle{Limitations of Existing Paradigms.}
The vanilla self-correction implementation~\cite{huang2023large} consistently degrades accuracy across both datasets ($-$1.52\% on \textsc{BIRD} and $-$1.29\% on \textsc{Spider}), confirming that black-box correction without explicit error modeling suffers from hallucination accumulation~\cite{qu2025share}.
Self-debugging~\cite{chen2024teach} achieves only modest gains (+1.40\% on \textsc{BIRD}), constrained by the low fraction of syntax errors in SQLs generated by backbone LLMs.

\subsection{RQ2: Task-specific Evaluation}
\label{section:rq2}

We evaluate the SQL error detection task and the text-to-SQL refinement task independently, and then analyze their joint effect on end-to-end performance.

\begin{table}[!t]

\centering
\caption{Results of SQL error detection based on GPT-4o generated SQLs.
$^{\diamondsuit}$ The detection mechanism of Refiner~\cite{wang2024macsql} is identical to self-debugging~\cite{chen2024teach}. 
$^{\clubsuit}$ The rules are in Section~\ref{section:detection}}
\label{table:detection}
\vspace{-3mm}
    \resizebox{\columnwidth}{!}{
    \begin{tabular}{lcccc}
        \toprule
        \textbf{Methods} & \textbf{Precision} & \textbf{Recall} & \textbf{D-Accuracy} & \textbf{D-F1}\\
        \midrule
        Self-detection~\cite{xia2024r3} & 47.70 & 29.10 & 54.63 & 36.15 \\
        \text{Self-correction}~\cite{pourreza2023dinsql} & 61.93 & 30.28 & 61.02 & 40.67 \\
        $\text{Refiner}^{\diamondsuit}$~\cite{wang2024macsql} & \textbf{100.0} & 2.95 & 57.17 & 5.74 \\
        SQLFixAgent~\cite{cen2025sqlfixagent} & 76.73 & \underline{70.61} & \underline{77.57} & \underline{73.54} \\
        \midrule
        \textbf{$\overleftarrow{\textsc{Rule}}^{\clubsuit}$} & \textbf{100.0} & 6.50 & 58.74 & 12.21 \\
        \textbf{\textsc{ErrorLLM}} & \underline{80.12} & \textbf{76.22} & \textbf{81.16} & \textbf{78.12} \\
        \bottomrule
    \end{tabular}
    }
\vspace{-5mm}
\end{table}

\stitle{SQL Error Detection Evaluation.} 
Table~\ref{table:detection} presents the SQL error detection results on GPT-4o generated SQLs from the \textsc{BIRD} development set.
\textsc{ErrorLLM} achieves the best overall performance with 81.16\% D-accuracy and 78.12\% D-F1.
Execution-based methods (Refiner and \textsc{Rule}) achieve perfect precision (100\%) but extremely low recall (2.95\% and 6.50\%, respectively), confirming that the vast majority of SQL errors are in semantic and do not trigger execution failures~\cite{liu2025nl2sqlbug}.
Notably, our \textsc{Rule} achieves the same flawless precision as execution-based self-debugging with higher recall.

\begin{figure}[!t]
    \centering
    \begin{tikzpicture}[scale=0.8]
    
    \def\ArrowTopY{102} 
    
    \def\ArrowRightX{600}

    \begin{axis}[
        height=5.8cm, 
        width=1.15\linewidth,
        xmin=-60, xmax=622, 
        ymin=20, ymax=106, 
        xtick={0, 100, 200, 300, 400, 500, 600},
        ytick={20, 30, 40, 50, 60, 70, 80, 90, 100},
        axis line style=ultra thick,
        xticklabel style={font=\small},
        yticklabel style={font=\small},
        xlabel={Number of True Positive Detections (TP)},
        ylabel={Fixed Rate (\%)},
        ylabel style={yshift=-0.3ex},
        xmajorgrids=true,
        ymajorgrids=true,
        grid style={dashed, gray!30},
        scatter style/.style={
            only marks, mark=*, mark size=4pt,
            mark options={draw=black!80, line width=1.2pt} 
        }
    ]
    
    \addplot[scatter style, fill=colOne] coordinates {
        (20, 60)    
        (205, 79.02)
    };

    \addplot[scatter style, fill=colTwo] coordinates {
        (20, 95)    
        (478, 36.61)
        (516, 33.53)
    };

    \addplot[scatter style, fill=colThree] coordinates {
        (516, 53.88)
    };

    \node[anchor=north west, font=\small, xshift=2pt, yshift=-2pt] at (axis cs:20, 95) {Refiner};
    \node[anchor=south west, font=\small, xshift=2pt, yshift=2pt] at (axis cs:20, 60) {Self-debugging};
    \node[anchor=south west, font=\small, xshift=2pt, yshift=2pt] at (axis cs:205, 79.02) {Self-correction};
    \node[anchor=south east, xshift=-4pt, yshift=2pt, font=\small] at (axis cs:478, 36.61) {SQLFixAgent};
    \node[anchor=south east, font=\small\bfseries, xshift=-4pt, yshift=2pt] at (axis cs:516, 53.88) {\textbf{\textsc{ErrorLLM} (Ours)}};
    \node[anchor=south, font=\small, yshift=-15pt] at (axis cs:516, 33.53) {$\text{Self-correction}^\diamondsuit$};

    \fill[gray!90, opacity=0.8] 
        (axis cs:30,  \ArrowTopY) --     
        (axis cs:560, \ArrowTopY + 1) --     
        (axis cs:580, \ArrowTopY) --    
        (axis cs:560, \ArrowTopY - 1) --  
        cycle;
    \node[below, font=\normalsize\bfseries, color=black!80] at (axis cs:305, \ArrowTopY) {Depth};

    \fill[gray!90, opacity=0.8] 
        (axis cs:\ArrowRightX, 25) --    
        (axis cs:\ArrowRightX + 5, 98) --    
        (axis cs:\ArrowRightX, 103) --  
        (axis cs:\ArrowRightX - 5, 98) --    
        cycle;

    \node[left, rotate=90, font=\normalsize\bfseries, color=black!80, anchor=south] at (axis cs:\ArrowRightX, 64) {Effectiveness};

    \end{axis}
    \end{tikzpicture}
    \vspace{-3mm}
    \caption{Comparison of text-to-SQL refinement methods by detection depth on GPT-4o generated SQLs from the \textsc{BIRD} development set. 
    $^\diamondsuit$ Self-correction~\cite{pourreza2023dinsql} uses the same number of TP as \textsc{ErrorLLM} for direct comparison.}
    \vspace{-5mm}
\label{figure:refinement}
\end{figure}
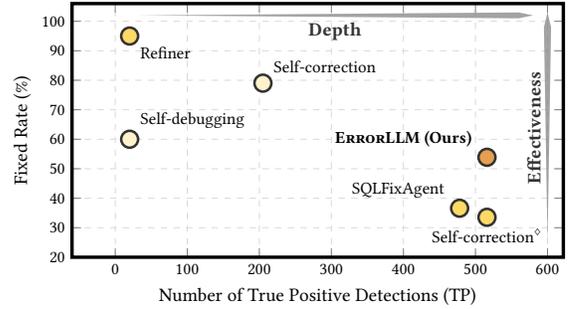

\stitle{Text-to-SQL Refinement Evaluation.}
We further compare the text-to-SQL refinement task in isolation, as shown in Figure~\ref{figure:refinement}.
An ideal refinement system should appear in the upper-right region, indicating both deep coverage and high efficiency.
Execution-based methods detect very few errors (around 20 samples), though Refiner achieves a high fixed rate (95\%) on the small number of detected samples.
Self-correction achieves a noteworthy fixed rate of 79.02\% on its 205 detected samples, demonstrating solid correction capability on shallow-detected errors.
\textsc{ErrorLLM} detects the most incorrect samples (516) while maintaining a substantially higher fixed rate (53.88\%) compared to both SQLFixAgent and the Self-correction$^\diamondsuit$ baseline at a similar detection scale.

\stitle{Joint Effect on Overall Performance.}
To understand how detection and refinement jointly determine end-to-end performance, we formalize the relationship between task-specific metrics and overall improvement.
Given a total of $A$ samples, the overall accuracy gain $\Delta\text{EX}$ from detection and refinement task can be decomposed as:
\begin{equation}
\label{equation:joint}
    \Delta\text{EX} = \frac{|\text{TP}| \times \text{FR} - |\text{FP}| \times \text{CR}}{A},
\end{equation}
where $|\text{TP}|$ is the number of truly incorrect SQLs detected (true positives), $|\text{FP}|$ is the number of correctly predicted SQLs falsely flagged (false positives), and FR and CR are as defined in Section~\ref{section:metrics}.

This decomposition reveals a fundamental tension in text-to-SQL refinement: improving $\Delta\text{EX}$ requires simultaneously maximizing detection recall (to increase $|\text{TP}|$) and precision (to minimize $|\text{FP}|$), while maintaining a high FR and low CR.
A critical insight from our experiments is that LLMs exhibit strong compliance with error-guided instructions during the text-to-SQL refinement task.
When provided with a false error signal, LLMs tend to modify the SQL regardless of its original correctness, resulting in an empirically high CR.
For example, Self-correction~\cite{pourreza2023dinsql} achieves a commendable FR of 79.02\% on its 205 detected errors, yet its overall $\Delta\text{EX}$ on \textsc{BIRD} is only $\uparrow$2.34\%.
This limited gain is attributable to the substantial false positives that lead to a large number of correct SQLs being corrupted, offsetting most of the gains from successful corrections.

\textsc{ErrorLLM} addresses both sides of Equation~\ref{equation:joint} effectively.
On the detection side, it achieves both the highest recall (76.22\%) and competitive precision (80.12\%);
On the refinement side, it maintains a FR of 53.88\% even at this large detection scale.
This dual advantage in detection quality and refinement efficiency yields the highest $\Delta\text{EX}$ across all methods, demonstrating that explicitly modeling SQL errors is essential for effective text-to-SQL refinement.

\subsection{RQ3: Fine-grained SQL Error Detection}
\label{section:rq3}

\begin{figure}[!t]
	\centering
	\includegraphics[width=\columnwidth]{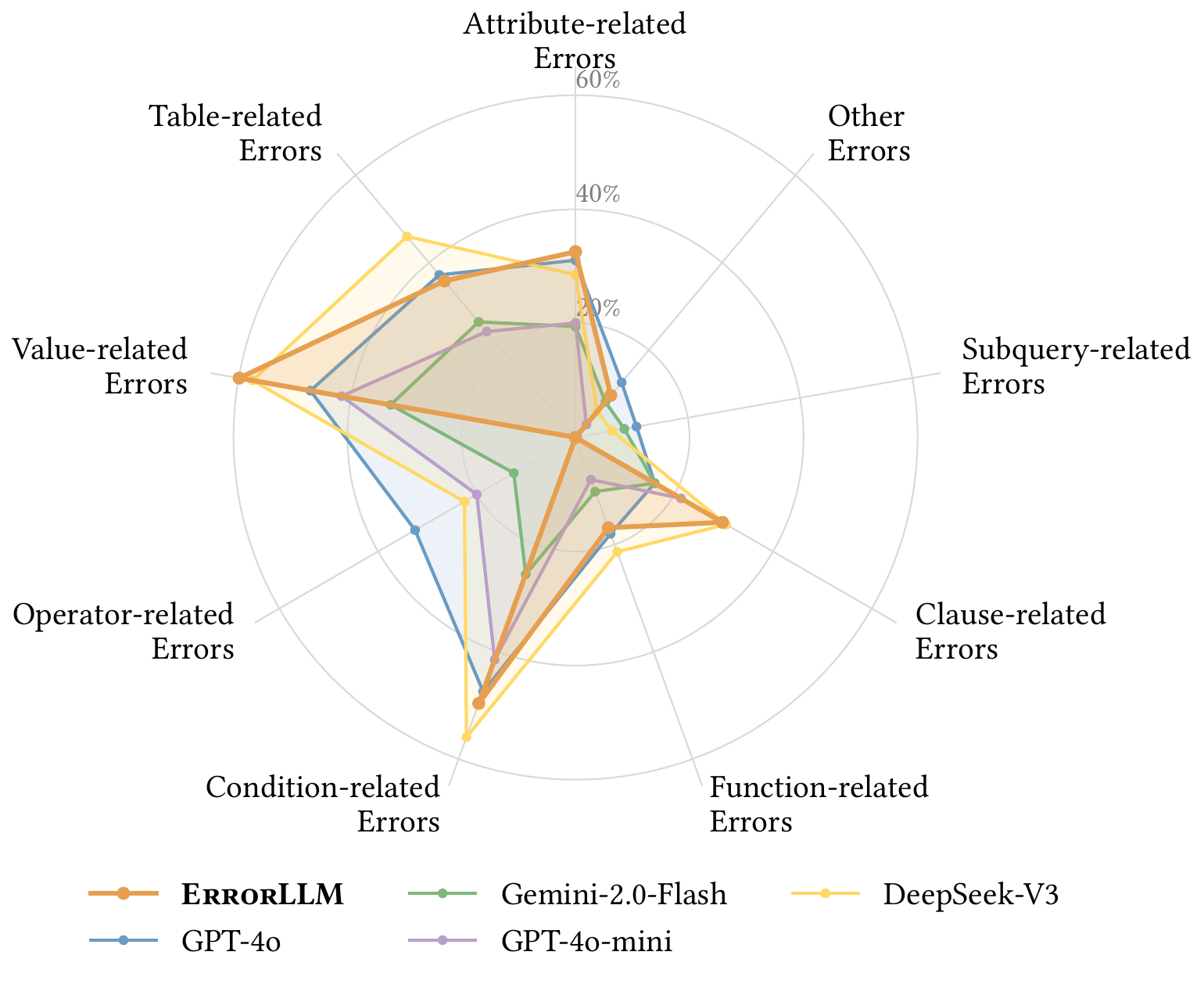}
\vspace{-3mm}
\caption{SQL error detection results of \textsc{ErrorLLM} and proprietary LLM baselines on \textsc{NL2SQL-Bugs}~\cite{liu2025nl2sqlbug}, evaluated per error category using type-specific accuracy (TSA).}
\vspace{-3mm}
\label{figure:categories}
\end{figure}

SQL error detection as modeled in \textsc{ErrorLLM} is inherently a fine-grained multi-label prediction task: it outputs a set of specific error tokens that map to $\Lambda$ (Table~\ref{table:taxonomy}) rather than a binary correct/incorrect decision.
This section evaluates whether such fine-grained predictions are accurate at the per-type level.
We adopt the \textsc{NL2SQL-Bugs}~\cite{liu2025nl2sqlbug} benchmark by mapping the two error taxonomies to enable a fair comparison.
\textsc{ErrorLLM} is compared with proprietary LLMs using the full \textsc{NL2SQL-Bugs} taxonomy, where these models have access to more fine-grained subcategory definitions.

As shown in Figure~\ref{figure:categories}, \textsc{ErrorLLM} achieves  comparable type-specific accuracy (TSA) on seven out of nine error categories, despite being a fine-tuned 7B model competing against proprietary LLMs with substantially larger parameter scales.
\textsc{ErrorLLM} reports zero TSA on ``subquery-related errors'' and ``operator-related errors'', as our current taxonomy $\Lambda$ does not define corresponding error types for these two categories.
This is a design limitation that can be studied in future work: the reserved error token slots ($[\textsc{Err}]_{13}$ to $[\textsc{Err}]_{N}$) in our vocabulary allow expansion to cover these categories, achieving seamless integration into the model architecture.
These results demonstrate that explicitly modeling SQL errors into \textsc{ErrorLLM}'s error tokens achieves fine-grained per-type detection quality.
This fine-grained detection capability largely empowers the downstream error-guided text-to-SQL refinement process: rather than receiving a vague ``the SQL is incorrect'' signal without further information, as discussed in Section~\ref{section:rq2}.

\subsection{RQ4: Ablation Study}
\label{section:rq4}

\begin{table}[!t]
\label{table:ablation}
\centering
\caption{Ablation studies on BIRD development set. 
$^\diamondsuit$~Structural representation is replaced by flat textual 
context.
$^\clubsuit$~question and schema nodes are fully connected.
$^\heartsuit$~Semantic initialization (Equation~\ref{equation:initialization}) is replaced by random initialization.
}
\label{table:ablation}

\vspace{-3mm}
    \resizebox{\columnwidth}{!}{
    \begin{tabular}{cl|cc}
        \toprule
        \textbf{Group} & \textbf{Variant} & \textbf{D-F1} & \textbf{EX} \\
        \midrule
        -- & \textbf{\textsc{ErrorLLM} (GPT-4o)} & -- & -- \\
        \midrule
        \multirow{3}{*}{\rotatebox{90}{\small\textbf{Structure}}}
        & w/o $\text{Question-schema structure}^\diamondsuit$ &  72.81 ($\downarrow$5.31) & 62.71 ($\downarrow$3.52) \\
        & w/o $\text{Abstract syntax tree}^\diamondsuit$ & 72.44 ($\downarrow$5.68) & 62.84 ($\downarrow$3.39) \\
        & w/o $\text{Linking structure}^\clubsuit$ & 73.53 ($\downarrow$4.59) & 63.62 ($\downarrow$2.61) \\
        \midrule
        \multirow{4}{*}{\rotatebox{90}{\small\textbf{Detection}}}
        & w/o Static detection & 76.25 ($\downarrow$1.87) & 65.97 ($\downarrow$0.26) \\
        & w/o LLM-based detection & 12.21 \textbf{($\downarrow$65.91)} & 58.41 \textbf{($\downarrow$7.82)} \\
        & w/o Constrained decoding & 73.72 ($\downarrow$4.40) & 64.15 ($\downarrow$2.08) \\
        & w/o $\text{Semantic initialization}^\heartsuit$ & 76.21 ($\downarrow$1.91) & 64.99 ($\downarrow$1.24) \\
        \midrule
        \multirow{2}{*}{\rotatebox{90}{\small\textbf{Data}}}
        & w/o Rule-based perturbation & 76.35 ($\downarrow$1.77) & 65.19 ($\downarrow$1.04) \\
        & w/o LLM-assisted injection & 67.55 (\underline{$\downarrow$10.57}) & 62.13 (\underline{$\downarrow$4.10}) \\ 
        \midrule
        \multirow{5}{*}{\rotatebox{90}{\small\textbf{Refinement}}}
        & w/o Error localization & -- & 62.19 ($\downarrow$4.04) \\
        & w/o Guideline templates & -- & 64.41 ($\downarrow$1.82) \\
        & w/o Subtree \& Subgraph & -- & 65.51 ($\downarrow$0.72) \\
        & w/o Few-shot examples & -- & 65.38 ($\downarrow$0.85) \\
        & w/o Priority ordering & -- & 64.93 ($\downarrow$1.30) \\
    \bottomrule
    \end{tabular}
}

\end{table}

To understand the contribution of each component, we conduct ablation studies on the BIRD dev set grouped by four functional roles.
Table~\ref{table:ablation} reports D-F1 for detection variants and EX for all variants; the full \textsc{ErrorLLM} achieves 78.12\% D-F1 and 66.23\% EX..

Among all variants, removing semantic detection causes the most severe degradation (D-F1 $\downarrow$65.91, EX $\downarrow$7.82), confirming that LLM-based error token prediction is the core mechanism of \textsc{ErrorLLM}.
The second most impactful factor is training data composition: removing LLM-assisted injection (D-F1 $\downarrow$10.57, EX $\downarrow$4.10) is far more damaging than removing rule-based perturbation (D-F1 $\downarrow$1.77, EX $\downarrow$1.04), demonstrating that learning from realistic LLM prediction behavior is essential for generalization.
Structural components (QSG, AST, linking structure) exhibit a notable amplification effect: their moderate D-F1 drops ($\downarrow$4.59 to $\downarrow$5.68) translate into disproportionately large EX drops ($\downarrow$2.61 to $\downarrow$3.52), because these representations serve a dual role in both error detection and the downstream refinement pipeline (error localization, subtree and subgraph extraction).
In the Refinement group, error localization is the most critical component ($\downarrow$4.04), followed by guideline templates ($\downarrow$1.82) and priority ordering ($\downarrow$1.30).
A consistent finding across all groups is that D-F1 degradation positively correlates with EX decrease, validating our framework's core hypothesis proposed in Section~\ref{section:rq2}

\section{Conclusion}

In this paper, we propose \textsc{ErrorLLM}, a framework that explicitly models text-to-SQL errors within a dedicated LLM for text-to-SQL refinement.
We extend the LLM's semantic space with dedicated error tokens, where each token corresponds to a categorized error type, and design a comprehensive SQL error detection approach that combines static detection with LLM-based semantic detection.
Guided by the detected errors, we further design an error-guided text-to-SQL refinement pipeline with error localization, error analysis, and priority-ordered correction to produce fixes.
Extensive experiments on \textsc{BIRD} and \textsc{Spider} benchmarks demonstrate that \textsc{ErrorLLM} achieves the most significant improvements over backbone initial generation, and is the only method that consistently improves both standard and strong backbones without corrupting originally correct SQLs.
Evaluation on the \textsc{NL2SQL-Bugs} benchmark further shows that the fine-grained error detection of \textsc{ErrorLLM} achieves competitive type-specific accuracy against proprietary LLMs.
Our analysis confirms that the quality of SQL error detection directly determines the effectiveness of text-to-SQL refinement, validating our core design of explicitly modeling SQL errors in the \textsc{ErrorLLM} as the foundation for effective text-to-SQL refinement.

\begin{acks}
This work was supported in part by a grant from the Innovation and Technology Commission of the Hong Kong Special Administrative Region, China (Project No. ITS/263/24FP), and in part by the National Natural Science Foundation of China (No. 62502008).
\end{acks}

\clearpage
\bibliographystyle{acm_reference}
\bibliography{reference}

\clearpage
\appendix

\section{Related Work}
\label{appendix:related}
\stitle{LLM-based Text-to-SQL.} LLMs~\cite{zhou2025taming} enable two paradigms: in-context learning (ICL) and fine-tuning. ICL methods such as DIN-SQL~\cite{pourreza2023dinsql}, DAIL-SQL~\cite{gao2023dailsql}, and CHESS~\cite{talaei2024chess} leverage prompting strategies, exemplar selection, and schema interaction. Fine-tuning approaches, including CodeS~\cite{li2024codes}, OmniSQL~\cite{li2025omnisql}, and SQL-R1~\cite{ma2025sql}, target complex queries and real-world domain-specific settings~\cite{hong2024knowledge}.
\stitle{Text-to-SQL Refinement.} Dedicated refinement frameworks such as MAC-SQL~\cite{wang2024macsql}, SQLFixAgent~\cite{cen2025sqlfixagent}, MAGIC~\cite{askari2025magic}, and SHARE~\cite{qu2025share} correct logical and semantic errors. Two paradigms dominate: self-debugging~\cite{chen2024teach}, limited by the small fraction of queries that trigger execution failures~\cite{liu2025nl2sqlbug}, and self-correction~\cite{huang2023large,pourreza2023dinsql}, which lacks error modeling and tends to corrupt already-correct SQLs~\cite{qu2025share}.
\begin{figure}[!t]
\begin{mdframed}[
    backgroundcolor=boxbg,
    linecolor=boxframe,
    linewidth=0.8pt,
    innerleftmargin=10pt,
    innerrightmargin=10pt,
    innertopmargin=8pt,
    innerbottommargin=8pt
]
\noindent\textbf{[User Question]} \hfill $q$

\smallskip
{\small\noindent\textit{``Find the names of students who have completed the Database course.''}\par}

\medskip
\noindent\textbf{[Ground-truth SQL]} \hfill $y$
\begin{lstlisting}[style=sqls]
SELECT s.name FROM student s
JOIN enrollment e ON s.id = e.student_id
WHERE e.status = (*@\hlg{\texttt{`Completed'}}@*);
\end{lstlisting}

\noindent\textbf{[Database Content]} \hfill $\operatorname{LookUp}(c,\,\mathcal{D})$

\smallskip
{\small\noindent\ttfamily
\hlg{`Active'}\ \ \hlg{`Completed'}\ \ \hlg{`Withdrawn'}\par}

\vspace{5pt}\noindent\rule{\linewidth}{0.3pt}\vspace{3pt}

\noindent\textbf{[Perturbation Operator]} \hfill $\operatorname{\textsc{Rule}}_7$

\smallskip
{\small
\noindent\textbf{Step 1.} Locate \texttt{Literal} $v^*$ in a comparison predicate.\\
\hspace*{1.5em}$\hookrightarrow$ \hlg{\texttt{`Completed'}}\\[2pt]
\noindent\textbf{Step 2.} Sample $v' \notin \operatorname{LookUp}(c, \mathcal{D})$ from an out-of-domain pool (morphological variant, near-synonym, format shift).\\
\hspace*{1.5em}$\hookrightarrow$ \hlr{\texttt{`Complete'}}\\[2pt]
\noindent\textbf{Step 3.} Substitute $v^* \mapsto v'$ in AST; flatten to $y'$, verified by $\operatorname{\textsc{Exec}}(y', \mathcal{D}) \neq \operatorname{\textsc{Exec}}(y, \mathcal{D})$. \quad\textcolor{softgreen}{\textbf{[confirmed erroneous]}}
\par}

\vspace{5pt}\noindent\rule{\linewidth}{0.3pt}\vspace{3pt}

\noindent\textbf{[Detection Rule]} \hfill $\overleftarrow{\operatorname{\textsc{Rule}}}_7$

\smallskip
{\small
\noindent\textbf{Step 1.} Traverse $\tau_{\hat{y}}$; locate \texttt{Literal} nodes in comparison predicates.\\
\hspace*{1.5em}$\hookrightarrow$ \hlr{\texttt{`Complete'}}\\[2pt]
\noindent\textbf{Step 2.} Resolve referenced column via QSS alignment.\\
\hspace*{1.5em}$\hookrightarrow$ $c = $ \texttt{enrollment.status}\\[2pt]
\noindent\textbf{Step 3.} Domain check $\hat{v} \stackrel{?}{\in} \operatorname{LookUp}(c, \mathcal{D})$.\\
\hspace*{1.5em}$\hookrightarrow$ \texttt{`Complete'} $\notin \operatorname{LookUp}(c, \mathcal{D})$, append \textbf{Value Error} to triggered error types $\{\hat{\lambda}\}_{\textsc{Rule}}$. \quad\textcolor{softred}{\textbf{[rule violation]}}
\par}
\end{mdframed}
\vspace{-3mm}
\caption{Paired construction for \textbf{Value Error} on a shared running example. The perturbation operator (top) injects an out-of-domain literal \texttt{`Complete'} to synthesize an erroneous SQL; the detection rule (bottom) flags the same literal.}
\label{figure:rules}
\vspace{-4mm}
\end{figure}

\section{Rule Illustration}
\label{appendix:rule}
Each error type $\lambda_i \in \Lambda$ is governed by a paired construction: a perturbation operator $\operatorname{\textsc{Rule}}_{\lambda_i}$ that synthesizes training data by injecting a targeted type of error into the ground-truth SQL, and a detection rule $\overleftarrow{\operatorname{\textsc{Rule}}}_{\lambda_i}$ that identifies the same class of error in predicted SQLs during inference.
We introduce an operator $\operatorname{LookUp}(c, \mathcal{D})$, which returns the set of all distinct values stored in column $c \in \mathcal{C}$ of database $\mathcal{D}$, and use it uniformly across both constructions.
We illustrate the paired design using \textbf{Value Error} as a representative example, with \texttt{enrollment.status} as the column $c$: the ground-truth predicate uses the enumerated value \texttt{`Completed'}, whereas a plausible LLM prediction uses the semantically adjacent but non-existent variant \texttt{`Complete'}, producing an empty result set.
As shown in Figure~\ref{figure:rules}, the perturbation operator replaces a valid literal with an out-of-domain value, execution-verified to confirm $\operatorname{\textsc{Exec}}(y', \mathcal{D}) \neq \operatorname{\textsc{Exec}}(y, \mathcal{D})$; the reversed detection rule traverses every \texttt{Literal} node in $\tau_{\hat{y}}$ and flags any literal absent from $\operatorname{LookUp}(c, \mathcal{D})$, achieving perfect precision by construction.

\section{Error Taxonomy}
\label{appendix:taxonomy}
Table~\ref{table:taxonomy} presents the error taxonomy $\Lambda$ in \textsc{ErrorLLM}, where each error type is mapped to a dedicated error token for explicit SQL error modeling. The taxonomy defines 12 error types inspired by recent studies on SQL error analysis~\cite{liu2025nl2sqlbug,yang2025hallucination,li2025deepeye,shute2024sql}. These error types frequently occur in both LLM and human SQL predictions~\cite{liu2025nl2sqlbug}, ensuring the generality of the detection task. To evaluate \textsc{ErrorLLM} on \textsc{NL2SQL-Bugs}~\cite{liu2025nl2sqlbug}, whose categories differ from ours, we map our taxonomy to theirs; categories without a corresponding type in $\Lambda$ (e.g., subquery- and operator-related errors) are left unmapped, detail mapping is provided in our repository for implementation.

\section{Efficiency Analysis}
\begin{table}[!t]
\centering
\caption{Efficiency analysis of \textsc{ErrorLLM} and baseline methods on the \textsc{BIRD} benchamrk.
$^\diamond$ The results are reproduced using official code; remaining baseline results are reported by the SHARE~\cite{qu2025share} under matching experimental settings.}
\vspace{-3mm}
\label{table:efficiency}
\setlength{\tabcolsep}{4pt}
\resizebox{\columnwidth}{!}{
\begin{tabular}{llcccc}
    \toprule
    & \multirow{2}{*}{\textbf{Method}} 
    & \multicolumn{2}{c}{\textbf{Open-source LLM}} 
    & \multicolumn{2}{c}{\textbf{Proprietary LLM}} \\
    \cmidrule(lr){3-4} \cmidrule(lr){5-6} & & \textbf{Input Tok.} & \textbf{Output Tok.} & \textbf{Input Tok.} & \textbf{Output Tok.} \\
    \midrule
    \multirow{4}{*}{\rotatebox{90}{\small\textbf{Training}}}
        & MAGIC~\cite{askari2025magic} & -- & -- & 4,838.6 & 2,085.2 \\
        & SHARE~\cite{qu2025share} & 2,308.8 & \underline{83.0} & 1,623.3 & 66.9 \\
        & \textbf{\textsc{ErrorLLM}} & 1,928.3 & \textbf{1.96} & 1,558.6  & 179.3  \\
    \midrule
    \midrule
    \multirow{6}{*}{\rotatebox{90}{\small\textbf{Inference}}}
        & Refiner~\cite{wang2024macsql} & -- & -- & 7,126.74 & 236.6 \\
        & Multiple-prompt~\cite{lee2025mcs} & -- & -- & 21,128.7 & 1,004.6 \\
        & $\text{SQLFixAgent}^\diamond$~\cite{cen2025sqlfixagent} & -- & -- & 4,839.1 & 345.2 \\
        & MAGIC~\cite{askari2025magic} & -- & -- & 8,245.2 & 1,738.0 \\
        & SHARE~\cite{qu2025share} & \textbf{1,731.2} & \underline{132.2} & \textbf{716.3} & \textbf{68.3} \\
        & \textbf{\textsc{ErrorLLM}} & \underline{1,739.4} & \textbf{1.45} & \underline{966.4} & \underline{213.5} \\
    \bottomrule
\end{tabular}
}
\vspace{-4mm}
\end{table}
Table~\ref{table:efficiency} compares token consumption against representative baselines. During training, \textsc{ErrorLLM}'s proprietary-LLM cost is a one-time overhead from LLM-assisted data synthesis alone, substantially lower than MAGIC~\cite{askari2025magic}, which relies on proprietary LLMs throughout. At inference, the most distinctive feature is the remarkably low open-source output token count (1.45 tokens/sample on average), a direct consequence of constrained decoding. The proprietary cost is bounded by detection recall: only samples flagged as erroneous trigger \textsc{LocLLM} and \textsc{RefLLM} calls, keeping average proprietary input tokens (966.4) competitive against methods that refine unconditionally. Overall, \textsc{ErrorLLM} attains the highest EX (Table~\ref{table:main}) while consuming substantially fewer proprietary tokens than most baselines. The token consumption calculation is implemented through the well-recognized titoken package\footnote{Available at \url{https://github.com/openai/tiktoken}} released by OpenAI.

\clearpage
\newpage
\begin{table*}[!t]
\centering
\caption{Error taxonomy where each error type is mapped to an error token in \textsc{ErrorLLM} for explicitly modeling SQL errors.}
\label{table:taxonomy}
\vspace{-3mm}
\resizebox{\textwidth}{!}{
\begin{tabularx}{\textwidth}{|l|l|X|X|}
\hline
\textbf{Error Token} & \textbf{Error Type} & \textbf{Description} & \textbf{Example} \\
\hline
\hline
$[\textsc{Err}]_{1}$ & Attribute Mismatch & 
The SQL selects incorrect attributes (columns) from the schema, indicating misalignment between the user question intent and attribute selection. & 
User question asks for ``student names'' but the corresponding SQL selects \lstinline[style=sqls]{student.id} instead of \
\lstinline[style=sqls]{student.name}. \\
\hline
$[\textsc{Err}]_{2}$ & Attribute Redundancy & 
The SQL includes unnecessary attributes (columns) not mentioned or implied in the user question. & 
User question asks ``list course names'' but the corresponding SQL selects \lstinline[style=sqls]{course.name, course.credits, course.department}, where 
\lstinline[style=sqls]{credits} and 
\lstinline[style=sqls]{department} are redundant/unnecessary. \\
\hline
$[\textsc{Err}]_{3}$ & Attribute Missing & 
The SQL fails to include essential attributes (columns) required by the user question. & 
User question asks for ``student names and their grades'' but the corresponding SQL only selects \lstinline[style=sqls]{student.name}, missing the 
\lstinline[style=sqls]{enrollment.grade} attribute. \\
\hline
$[\textsc{Err}]_{4}$ & Table Mismatch & 
The SQL references incorrect tables from the database schema. & 
User question asks about ``course enrollments'' but SQL queries the \lstinline[style=sqls]{attendance} table instead of the 
\lstinline[style=sqls]{enrollment} table. \\
\hline
$[\textsc{Err}]_{5}$ & Table Redundancy & 
The SQL joins unnecessary tables not required by the query logic. & 
User question asks ``list all students'' but SQL joins \lstinline[style=sqls]{student}, \lstinline[style=sqls]{enrollment}, and \lstinline[style=sqls]{course} tables when only 
\lstinline[style=sqls]{student} is needed. \\
\hline
$[\textsc{Err}]_{6}$ & Table Missing & 
The SQL omits required tables needed to answer the user question. & 
User question asks ``find student names and their course grades'' but the corresponding SQL only queries the \lstinline[style=sqls]{student} table without joining 
\lstinline[style=sqls]{enrollment}. \\
\hline
$[\textsc{Err}]_{7}$ & Value Error & 
The SQL uses incorrect values or wrong data formats in conditions. & 
User question asks for ``courses after January 1, 2023'' but SQL uses \lstinline[style=sqls]|start_date > '01/01/23'| instead of 
\lstinline[style=sqls]|'2023-01-01'|, or uses 
\lstinline[style=sqls]|enrollment > '30'| instead of 
\lstinline[style=sqls]{30}. \\
\hline
$[\textsc{Err}]_{8}$ & Condition Missing & 
The SQL fails to include necessary filtering conditions stated or implied in the user question. & 
User question asks ``students who completed the Database course'' but SQL omits \lstinline[style=sqls]|WHERE course.name = 'Database'| and the implicit 
\lstinline[style=sqls]|grade IS NOT NULL| check. \\
\hline
$[\textsc{Err}]_{9}$ & Condition Error & 
The SQL contains incorrect or redundant conditions that deviate from the user question intent. & 
User question asks for ``Grade 3 students with Math OR English scores above 90'' but SQL uses \lstinline[style=sqls]|Grade = 3 AND Math > 90 OR English > 90| without proper parentheses. \\
\hline
$[\textsc{Err}]_{10}$ & Function Error & 
The SQL misuses aggregate, date/time, string, or conditional functions. & 
User question asks ``average grade above 85 for each course'' but the corresponding SQL places \lstinline[style=sqls]{AVG(grade)} in \lstinline[style=sqls]{WHERE} instead of \lstinline[style=sqls]{HAVING}, or omits 
\lstinline[style=sqls]{ROUND(AVG(grade), 2)}. \\
\hline
$[\textsc{Err}]_{11}$ & Clause Error & 
The SQL incorrectly includes or omits essential clauses like GROUP BY, ORDER BY, or HAVING. & 
User question asks to ``count students per department'' but the corresponding SQL omits the \lstinline[style=sqls]{GROUP BY department} clause, or includes unnecessary 
\lstinline[style=sqls]{ORDER BY}. \\
\hline
$[\textsc{Err}]_{12}$ & Modifier Error & 
The SQL incorrectly uses or omits modifiers like DISTINCT, ASC, or DESC. & 
User question asks for ``unique course names in descending order'' but SQL omits \lstinline[style=sqls]{DISTINCT} or uses 
\lstinline[style=sqls]{ORDER BY name ASC} instead of 
\lstinline[style=sqls]{DESC}. \\
\hline
$[\textsc{Err}]_{13}$ \text{--} $[\textsc{Err}]_{N}$ & Reserved Slots & 
Pre-extended error token slots ($N > |\Lambda|$) are reserved for future error types to ensure model scalability. & 
These slots remain inactive until new error categories are added to the taxonomy $\Lambda$. \\
\hline
$[\textsc{Err}]_{\varnothing}$ & No Error & 
The SQL query is correct and accurately represents the user question intent. & 
The SQL produces the expected execution results (correct results) for the given user question and database schema. \\
\hline
\end{tabularx}
}
\end{table*}

\end{document}